\documentclass[fleqn,10pt]{wlscirep}
\usepackage[utf8]{inputenc}
\usepackage[T1]{fontenc}
\usepackage{multirow}
\usepackage{tikz}
\usetikzlibrary{decorations.pathreplacing,calc}
\usepackage{braket}
\usepackage{algorithm}
\usepackage{algpseudocode}
\usepackage{array}
\usepackage{longtable}
\usepackage{subcaption}
\usepackage{cleveref}
\usepackage{appendix}
\usepackage{bm}
\usepackage[mathcal]{euscript}

\newtheorem{Definition}{Definition}

\newcommand{\blfootnote}[1]{%
  \begingroup
  \renewcommand{\thefootnote}{}\footnote{#1}%
  \addtocounter{footnote}{-1}%
  \endgroup
}

\title{TabularQGAN: A quantum generative model for tabular data synthesis}

\author[* +1,3]{Pallavi Bhardwaj}
\author[+2,3]{Caitlin Jones}
\author[1,4,5]{Lasse Dierich}
\author[3,6]{Aleksandar~Vučković}
\affil[1]{SAP SE, Walldorf, Germany }
\affil[2]{BASF Digital Solutions, Ludwigshafen, Germany}

\affil[3]{QUTAC, Quantum Technology and Application Consortium, Germany}
\affil[4]{Technical University of Munich, CIT, Garching, Germany }
\affil[5]{Ludwig-Maximilians-Universität München, Munich, Germany }

\affil[6]{Merck KGaA, Darmstadt, Germany}

\affil[*]{pallavi.bhardwaj@sap.com}

\affil[+]{these authors contributed equally to this work}

\begin{abstract}
In this paper, we introduce a novel quantum generative model for synthesizing tabular data. Synthetic data is valuable in scenarios where real-world data is scarce or private, as it can be used to augment or replace existing datasets. As enterprise data is predominantly tabular and heterogeneous, often consisting of both categorical and numerical features,  this task is relevant across various industries such as healthcare, finance, and software. Existing quantum generative models are designed for homogeneous data; we seek to fill this gap by proposing a quantum generative adversarial network architecture with flexible data encoding and a novel quantum circuit ansatz for effectively modeling tabular data. The proposed approach is tested on the MIMIC-III healthcare and Adult Census datasets, with extensive benchmarking against leading classical models, CTGAN, CopulaGAN, VAE-GMM, and an LLM-based approach using the be-GReaT framework for tabular data synthesis. 
We evaluated our model as a proof-of-concept on reduced feature subsets using a noiseless statevector simulator on classical hardware. Our simulations show that, for the MIMIC-III dataset, our quantum model achieves competitive, and in some cases,
leading performance with respect to an overall similarity score used in the open-source Python library SDMetrics. Additionally, we evaluate the generalization capabilities of the models using two custom-designed metrics that demonstrate the ability of the proposed quantum model to generate useful and novel tabular samples.

\end{abstract}
\begin{document}

\flushbottom
\maketitle
\thispagestyle{empty}
\blfootnote{This is the version of the manuscript accepted for publication in \emph{Scientific Reports}. Please cite the published version: Bhardwaj, P., Jones, C., Dierich, L. \& Vučković, A. TabularQGAN: a quantum generative model for tabular data synthesis. \textit{Sci. Rep.} \textbf{16}, 23555 (2026). \url{https://doi.org/10.1038/s41598-026-62434-1}}

 % ------------------------------------------------------------------
% Introduction(1-2 and half page)
%	1. Overview of the problem 
%		a. Why tabular data and its relevance
%		b. Why quantum computing for this problem
%	2. Current research in this area
%	3. Why have we chosen QGAN for tabular data
%		a. Start with QGAN and about quantum computing(QGAN 2 papers from Loyd and                          Killoran )
%		b. Specifics of the architecture + ansatz(similar to point 4 though)
%	4. Purpose of our Algorithm and issues it addresses
%   5. Brief overview of the paper
%
% --------------   ----------------------------------------------------
% Merge introduction and related work
\section{Introduction}
Recent progress in quantum computing research for both hardware \cite{acharya2024quantum, bluvstein_logical_2023} and algorithmic \cite{kim_evidence_2023, jordan2024optimization} aspects has been promising. It remains to be shown if quantum computers are universally faster, more energy efficient, or otherwise more useful than classical computers for a task that is universal to an entire set of problems, apart from areas where such an advantage has been successfully proven: factorization \cite{doi:10.1137/S0097539795293172}, unstructured search \cite{grover_1996}, and quantum simulation~\cite{qpe}.

In this work, we investigate quantum machine learning (QML) models, a class of machine learning models which incorporates quantum computing into a large variety of machine learning architectures and tasks including neural networks \cite{schuld2018supervised}, reinforcement learning \cite{garcia2019quantum}, transformers \cite{cherrat2022quantum, gao2023fast}, image classification \cite{senokosov2024quantum}, and, the subject of this work, unsupervised generative models. References \cite{Sweke2021quantumversus} and \cite{wang2024comprehensive} provide recent overview of the field. As of now, there is no demonstration of QML providing a robust and repeatable advantage over classical methods for practically useful problems \cite{bowles2024better}. However, prior work has investigated whether quantum models can represent certain distributions compactly or achieve competitive performance with fewer trainable parameters in selected settings, and has found positive results. Therefore, the expressivity of quantum models may achieve the same or better performance with a large reduction in the number of parameters required for certain tasks, such as generative learning \cite{abbas2021power}. Additionally, recent work has shown that some quantum models can perform generative learning tasks efficiently that are classically hard (in terms of the number of steps and data samples) \cite{huang2025generative}.  
Generative QML methods such as Quantum Circuit Born Machines (QCBM)~\cite{e20080583, qcbm_quantum} and Quantum GANs (QGAN)~\cite{PhysRevA.98.012324, lloyd_quantum_2018, zhu2019}, have demonstrated comparable training performance to classical models, requiring fewer parameters \cite{riofrio2023performance}. QML may be particularly well suited to generative tasks, as the fundamental task of a quantum computer is to produce a probability distribution that is sampled from, which is also what is required for a generative task \cite{gao2018}.  

Generative models are useful in cases whereby amplifying a sample population with generated samples can lead to improved statistics of rare events \cite{zheng2021generative} (to be used in anomaly detection, fraud identification and market simulations), to improve generative design pipelines (for drug discovery and personalized assistants \cite{zeng2022deep}) and alleviate data privacy concerns by sharing synthetic instead of confidential data \cite{keshta2021security, mayer2020privacy}.
 
The majority of generative QML research has been on homogeneous data, e.g., image and text data, but business-relevant data is often heterogeneous tabular data, i.e., data that has numeric, categorical, as well as binary features. A prominent example is electronic health records (EHR), which are collections of heterogeneous patient data \cite{xiao_opportunities_2018, ghosheh_review_2022} and are subject to particularly strong privacy constraints. Other examples include human resources-related data and chemical structures.

Previous work investigated the use of quantum kernel models in a classification setting of EHR data~\cite{krunic_quantum_2022}, as well as the use of classical GANs to model EHR data (medGAN) with a reduction of continuous features to a discrete latent space via autoencoding~\cite{choi_generating_2017}. In addition, the generation of heterogeneous time series data has been investigated, employing variational autoencoders to map the data to and from a smaller latent space for training \cite{li_generating_2023}. 

In this work, we introduce a novel method for generating tabular data with a QGAN using a custom ansatz (quantum circuit) that does not require additional autoencoding. The model is an adaptation of the model presented in \cite{riofrio2023performance}. The architecture makes use of a new approach to model one-hot vectors representing the exclusive categorical features, and is, by construction, well suited to represent numerical features. We perform hyperparameter optimization and benchmark against classical models \cite{xu2019modeling, sdv} on selected feature subsets of two datasets, MIMIC-III \cite{johnson2016mimic, mimic_db} and Adult Census \cite{adult_census}. We find that the best-found configurations of the quantum tabular model had higher values of the overall similarity score compared to our classical benchmarks (a CTGAN, CopulaGAN, VAE-GMM, be-GReaT LLM) for the MIMIC-III dataset and all but one classical model (the VAE-GMM) for the Adult Census data. 

 In \cref{sec:variation_qc} we introduce the mathematical framework of quantum generative models and our approach to modeling one-hot vectors, in \cref{sec:encoding}, we describe how data is encoded into the quantum circuit, in \cref{sec:Quantum_Generator} we outline the specific architecture of our quantum generative model and how it is trained, followed by an analysis of the resources required from a quantum computer, in \cref{sec:resource_analysis}. In \cref{sec:training_qgan} the training procedure for the quantum model is described and in \cref{sec:eval_benchmarking}, the benchmarking and evaluation metrics are outlined. In \cref{sec:datasets} details of the datasets are given, and in \cref{sec:hyper_param_opt}, the hyperparameter optimization procedure is outlined. The experimental configuration and numerical results are presented in \cref{sec:numerical_results}. Finally, a discussion of the results, the limitations of our approach, and an outlook is presented in \cref{sec:conclusion}. 

\section{Methodology} \label{sec:methodology}

% Theory/Methodology(3-4 pages)

% 	1. Theoretical foundation of our model(description of our ansatz)
% 		a. Briefly introduce QGAN
% 		b. Details about our ansatz
%            i. registers in the circuit
% 			i.   Qubits encoding with example
% 			ii.  Circuit division and Specific gates and its relevance
% 			iii. How scaling effects number of gates and qubits
%       c. Preposition/Definition/Theorem
%            i.   Addressing Barren (move to discussion)plateaus(theoretical side, refer to literature), refer lie algebra paper for classical                          simulatibility of ansatz
%            ii.  Reduction in the size of Hilbert space by given rotation + sample                    space for continuous variable ansatz(check existing literature)
%            iii. Entanglement and relationship ( Fisher information for measuring entanglement of states)       
%        d. Generalization metrics
% 	2. optional Algorithm in pseud-code format(optional but preferred by Pallavi)
% 	3. Hyperparameter optimization(caitlin)
%   4. Classical SOTA models and classical vanilla GAN
%   5. Evaluation metrics and evaluation with XGboost classifier

% Briefly introduce QGAN
\subsection{Quantum Generative Models and Variational Quantum Circuits} \label{sec:variation_qc}

Any generative model is expected to possess two key capabilities: learning the underlying data distribution of the training data, and generalizing beyond it to generate novel samples. In this work, we primarily focus on the ability to learn the probability distribution, but we also consider metrics to measure generalization. We propose a quantum–classical generative adversarial network (QGAN), which has a structure similar to that of a conventional GAN, with the generator implemented using a variational quantum circuit (VQC) \cite{vqa}, and the discriminator realized using a classical neural network (Figure~\ref{fig:Schematic-Diagram}).

A VQC is a sequence of parametrized unitary matrices (\textit{gates}) $U_m(\theta_m)~\in~\mathbb{C}^{2^n \times 2^n}$ that prepares a \textit{quantum state} 
\begin{equation}
\ket{\Psi(\{\theta_m\}_m)} = \prod_{m} U_m(\theta_m)\ket{\Psi_{0}}~,
\end{equation}
starting from an $n$-qubit initial quantum state $\ket{\Psi_{0}}$ of the finite-dimensional \textit{Hilbert space} $\mathbb{C}^{2^n}$, where $U_m$ is the gate at index $m$ and $\theta_m$ is its scalar parameter. The \textit{computational basis}
\begin{equation}
\left\{\left|i_1 i_2 \dots i_n\right\rangle\right\}_{i_1, \ldots, i_n \in\{0,1\}} := \left\{\left|i_1\right\rangle \otimes\left|i_2\right\rangle \otimes \cdots\left|i_n\right\rangle\right\}_{i_1, \ldots, i_n \in\{0,1\}}
\end{equation}
is an orthonormal basis for this space and $\left \langle i_1 i_2 \dots i_n\right|$ is the conjugate transpose of the vector $\left|i_1 i_2 \dots i_n\right \rangle$. Quantum mechanics allows the quantum state $\ket{\Psi(\{\theta_m\}_m)}$ to be a superposition of multiple basis states, which collapses with a probability of
\begin{equation}
p(i_1 i_2 \dots i_n) =|\langle i_1 i_2 \dots i_n \ket{\Psi(\{\theta_m\}_m)}|^2
\end{equation}
into the state $\left|i_1 i_2 \dots i_n\right \rangle$ upon measurement. To employ the VQC as a generator $G$, we encode the rows of a given tabular training set into bitstrings $i_1 i_2 \dots i_n$ and aim to find a set of parameters $\{\theta_m\}_m$ such that the probability distribution $p(i_1 i_2 \dots i_n)$ over all bitstrings approximates the underlying probability distribution of the training data. In contrast, the role of the classical discriminator $D: \{0, 1\}^{n} \rightarrow (0, 1)$ is to distinguish whether a bitstring $x$ is a genuine (0) or synthetic sample (1). In our experimental evaluation, we consider tabular datasets comprising both numerical and categorical features. Numerical features are modeled using the discrete circuit architecture from \cite{riofrio2023performance}, while categorical variables are encoded through the application of Givens rotations.

As demonstrated in \cite{Arrazola_2022}, controlled single-excitation gates implemented as Givens rotations form a universal gate set for particle-conserving unitaries in quantum chemistry. Givens rotations are unitary transformations within a designated subspace of a larger Hilbert space, and we adapt these rotations to preserve the one-hot encoding intrinsic to categorical features. In quantum systems with a fixed excitation number, these rotations facilitate transitions only among basis states that maintain the total number of excitations. For example, in a system comprising $n$ qubits with exactly $k$ excitations, the relevant subspace is spanned by all states in which exactly $k$ qubits are in the excited state $\ket{1}$ and the remaining $n-k$ qubits are in the ground state $\ket{0}$. The dimensionality of this subspace is \(d=\binom{n}{k}\).

\begin{Definition}

[Particle Conserving Unitaries and Givens Rotations] \label{prop1}

Let $\mathcal{H} = (\mathbb{C}^{2})^{\otimes n}$ denote the Hilbert space of an $n$-qubit system, and let $N = \sum_{i=1}^n |1\rangle\langle 1|_i$ denote the total excitation number operator that counts the number of qubits in the excited state $\lvert 1 \rangle$. A unitary operator $U \in U(2^n)$ is defined as particle-conserving if it satisfies the commutation relation with the total excitation operator \cite{marvian2024theory,lacroix2022symmetry}
\begin{equation}[U, N] = UN - NU = 0~.\end{equation} This commutator implies that measuring or counting the total excitations before or after the application of the unitary $U$ yields identical results, and the operator $U$ cannot create or destroy excitations, it can only redistribute or swap them among the qubits. Consequently, if the system is initialized in the $k$-excitation subspace $\mathcal{H}_k \subset \mathcal{H}$, the span of all computational basis states containing exactly $k$ qubits in state $\lvert 1 \rangle$ and $n-k$ qubits in state $\lvert 0 \rangle$, the unitary $U$ leaves this subspace invariant. Following the matrix decomposition strategies established by Reck et al. \cite{reck1994experimental} and Clements et al. \cite{clements2016optimal}, any such particle-conserving unitary transformation $U$ acting on a finite-dimensional excitation space can be exactly decomposed into a product of two-level unitary operations known as Givens rotations:\begin{equation}U(\theta,\varphi) = \prod_{m} G_{i_m, j_m}(\theta_m, \varphi_m)\end{equation}where each individual Givens rotation $G_{ij}(\theta, \varphi)$ acts exclusively on the single-excitation subspace of a specific pair of qubits $i$ and $j$ (spanning the basis states $\lvert 0_i 1_j \rangle$ and $\lvert 1_i 0_j \rangle$). In this restricted two-dimensional basis, the operator takes the matrix form
\begin{equation}
    \label{eq:given_product}
    G_{ij}(\theta,\varphi) = \begin{bmatrix} 
    \cos\theta & -e^{i\varphi} \sin\theta \\
    e^{-i\varphi} \sin\theta & \cos\theta
    \end{bmatrix}~.
\end{equation}
Each $G_{ij}(\theta, \varphi)$ only alters the probability amplitudes of the basis states $\lvert x_i \rangle$ and $\lvert x_j \rangle$ without affecting any other state, ensuring that the overall transformation remains strictly within $\mathcal{H}_k$.

\end{Definition}

Definition \ref{prop1} is particularly useful for encoding categorical features in tabular datasets. Let $\mathcal{C}$ be a categorical feature with $n$ distinct categories and let $\mathcal{H} = (\mathbb{C}^2)^{\otimes n}$ be the Hilbert space of $n$ qubits. The one-hot encoding subspace $\mathcal{H}_{\text{OH}} \subset \mathcal{H}$ is the 1-excitation manifold (Hamming weight $k=1$) with dimension $d = \binom{n}{1} = n$. Any unitary $U$ that preserves the one-hot validity of the categorical feature (i.e., satisfies $[U, N]=0$) can be exactly represented as a product of $M \leq \frac{n(n-1)}{2}$ Givens rotations $G_{ij}(\theta, \varphi)$, following the universal unitary decomposition theorems established by Reck et al. \cite{reck1994experimental} and Clements et al. \cite{clements2016optimal}. These rotations act strictly within the span of the basis states $\{ \lvert 100\dots \rangle, \lvert 010\dots \rangle, \dots, \lvert 000\dots 1 \rangle \}$, ensuring that the generated synthetic data never violates the categorical constraints. Consider a categorical feature that contains $n=9$ distinct categories. Under a standard qubit mapping, let each category correspond to a basis state in $\mathcal{H} = (\mathbb{C}^2)^{\otimes 9}$. By restricting the generator to the one-hot manifold $\mathcal{H}_{\text{OH}}$ (the $k=1$ excitation subspace), we ensure that any synthetic sample satisfies the categorical constraint $\langle \psi | N | \psi \rangle = 1$ by construction. This reduces the effective dimensionality of the optimization manifold from $2^9$ to $\binom{9}{1} = 9$. By applying the decomposition $U = \prod_{m} G_{i_m j_m}(\theta_m, \varphi_m)$, we see potential benefits for generative modeling. First, it enables implicit constraint satisfaction by construction by restricting the quantum circuit not to generate any invalid states (e.g., multi-class assignments), as $G_{ij}$ is a symmetry-preserving operation that ensures the state never leaves $\mathcal{H}_k$. Furthermore, we conjecture that this approach may enhance optimization stability by constraining the search space to the relevant symmetry sector, where the optimizer navigates a manifold of substantially lower volume compared to the unconstrained Hilbert space. In the case of $n=9$ categories, the effective dimensionality is reduced from $2^9 = 512$ to $n=9$. By concentrating the gradient signal within a low-dimensional, physically relevant subspace, we also conjecture that the ansatz has the possibility to alleviate the problem of the exponential vanishing of gradients. The vanishing of gradients due to the high dimensionality of the Hilbert space is also known as "barren plateaus", and because of the dimensionality reduction due to the restriction to a subspace, there is potential to avoid this issue in some cases \cite{schatzki2024theoretical}, although we do not provide explicit proof or numerical studies in this work. For $k=1$, where the subspace dimension remains equal to the number of categories $n$, the reduction in the parameter space can be defined as the ratio of the dimensions of the full Hilbert space to that of the one-hot subspace:
\begin{equation}
    \text{RF}(n) = \frac{\dim(\mathcal{H}_{\text{full}})}{\dim(\mathcal{H}_{OH})} = \frac{2^n}{n}
\end{equation}
We suggest that this scaling may help ensure that the QGAN remains parameter-efficient even as the cardinality of the categorical features expands. Recent work by Ma et al. \cite{ma2024parameterefficientquasiorthogonalfinetuning} demonstrates that orthogonal fine-tuning via Givens rotations significantly reduces parameter complexity while retaining performance.

\subsection{Encoding}\label{sec:encoding}
In this section, we will describe how the tabular data is encoded into quantum states via basis encoding. Each data sample is mapped to a bitstring of length $n$, which is split into the \textit{numerical register}, containing ordered variables, and the \textit{categorical register} containing unordered variables. All numerical variables \(x\) are partitioned into \(2^N\) equal-width bins, where \(N\) is the number of qubits allocated to $x$ (the qubit budget). The index \(i \in \{0,\dots,2^N-1\}\) of the respective bin is represented by a computational-basis state,
\begin{equation}
\bigl\lvert b_{N-1} b_{N-2} \dots b_{0} \bigr\rangle~,
\end{equation} 
where \((b_{N-1}\dots b_{0})\) is the \(N\)-bit binary expansion of index \(i\).

Categorical features with multiple classes (\(c>2\)) are one-hot encoded using a dedicated \(c\)-qubit subregister and binary features (\(c=2\)) are encoded either as Boolean with one qubit (\(\lvert0\rangle\) vs.\ \(\lvert1\rangle\)) or one-hot with two qubits (\(\lvert10\rangle\) vs.\ \(\lvert01\rangle\)). This dual-encoding strategy yields two distinct circuit topologies, as shown in Figure \ref{fig:circuits}. The full input register is obtained by concatenating the subregisters for each feature. If a feature \(f\) uses \(N_f\) qubits, then an entire record is represented as
\begin{equation}
    \bigl\lvert \underbrace{b^{(1)}_{N_1-1}\dots b^{(1)}_{0}}_{\text{feature 1}} \;\big\|\;\underbrace{b^{(2)}_{N_2-1}\dots b^{(2)}_{0}}_{\text{feature 2}} \;\big\|\;\cdots \bigr\rangle
\end{equation}
in the computational basis. An explicit example of this encoding can be found in Appendix~\ref{sec:encoding_example}. We also tested a binary encoding of features as an alternative to one-hot encoding: Each combination of feature values can be assigned an index, allowing joint features to be mapped onto a single numerical register. This approach is described in more detail in Appendix~\ref{sec:unique-row-encoding-example}, but it performed significantly worse than the one-hot encoding.
%Circuit Structure

\subsection{Quantum Generator} \label{sec:Quantum_Generator}
\begin{figure}[h]
\label{circuit_design}
\begin{minipage}[t]{0.49\textwidth}
       \begin{tikzpicture}
            \tiny
            \node (fig1) at (0,0)
                {\includegraphics[height=4.85015 cm, clip, trim=0 0 4.3cm 0]{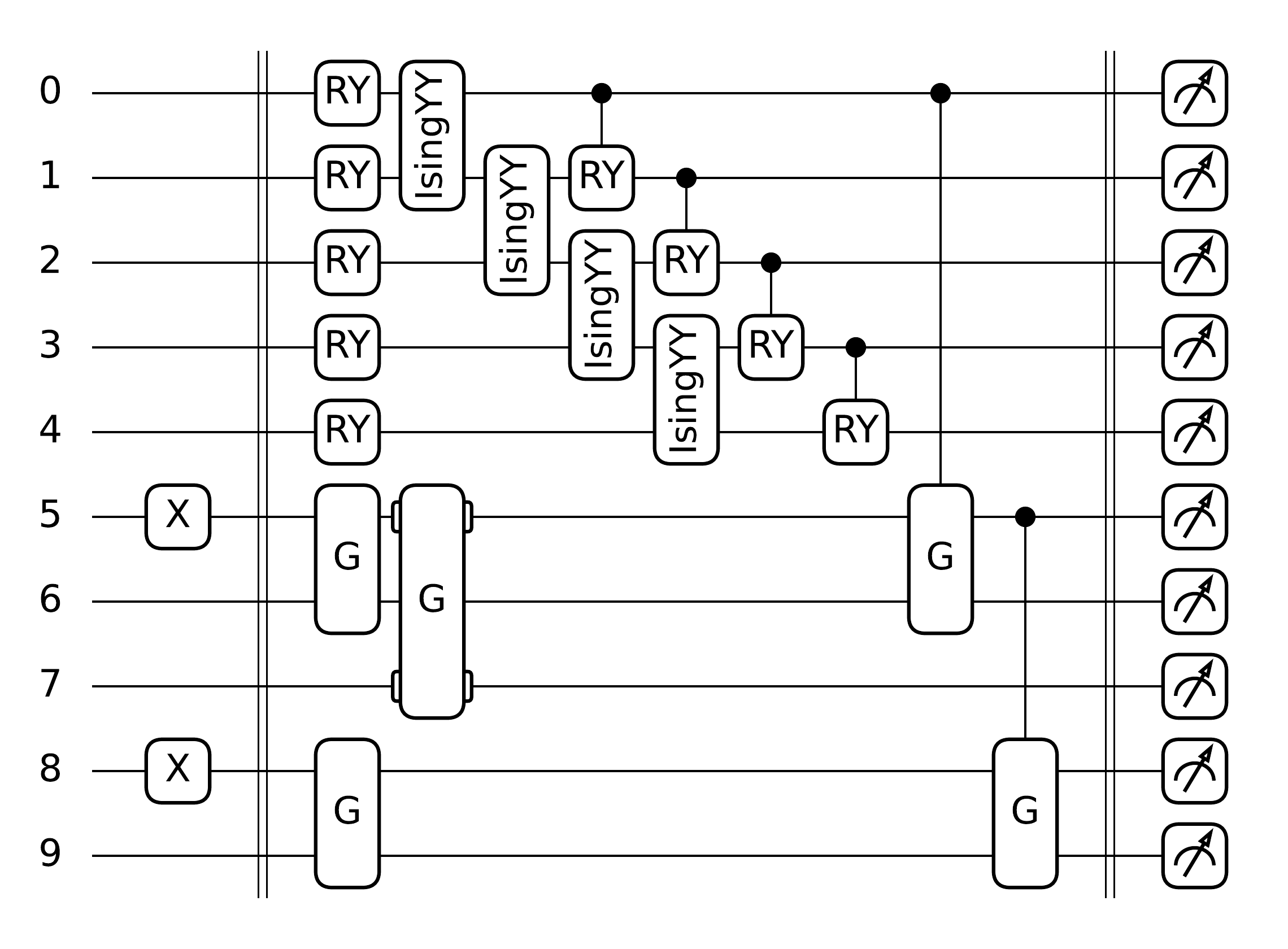}};
            %\node (fig1) at (3.9,0)
            %    {\includegraphics[height=4.85015 cm, clip, trim=32.5cm 0 0 0]{figures/non-bool.pdf}};
        
            % Insert \dots 10 times vertically aligned
            \foreach \i in {1,2,...,10} {
                \node at (3.12, -2.38+0.434*\i) {\tiny \dots};
            }
            % The layers
            \draw[decorate,decoration={brace,amplitude=4pt}] (-1.65, -2.29) -- (-2.45, -2.28) node[midway, below,yshift=-4pt,xshift=4pt]{\tiny initial excitation};
            \draw[decorate,decoration={brace,amplitude=4pt}] (2.76, -2.28) -- (-1.53, -2.29) node[midway, below,yshift=-4pt,]{\tiny model layer};
            % The register sizes
            \draw[decorate,decoration={brace,amplitude=4pt}] (-2.8, 0.09) -- (-2.8, 2.1) node[midway, below,xshift=-9pt, yshift=4.5pt]{\tiny n5};
            \draw[decorate,decoration={brace,amplitude=4pt}] (-2.8, -1.2) -- (-2.8, -0.03) node[midway, below,xshift=-9pt, yshift=4.5pt]{\tiny c3};
            \draw[decorate,decoration={brace,amplitude=4pt}] (-2.8, -2.1) -- (-2.8, -1.32) node[midway, below,xshift=-9pt, yshift=4.5pt]{\tiny c2};
        \end{tikzpicture}
        \caption*{(a) Non-Boolean design}
\label{fig:non-bool} 
\end{minipage}
\hfill
\begin{minipage}[t]{0.49\textwidth}
       \begin{tikzpicture}
            \tiny
            \node (fig1) at (0,0.335)
                {\includegraphics[height=4.225 cm, clip, trim=0 3.8cm 4.3cm 0]{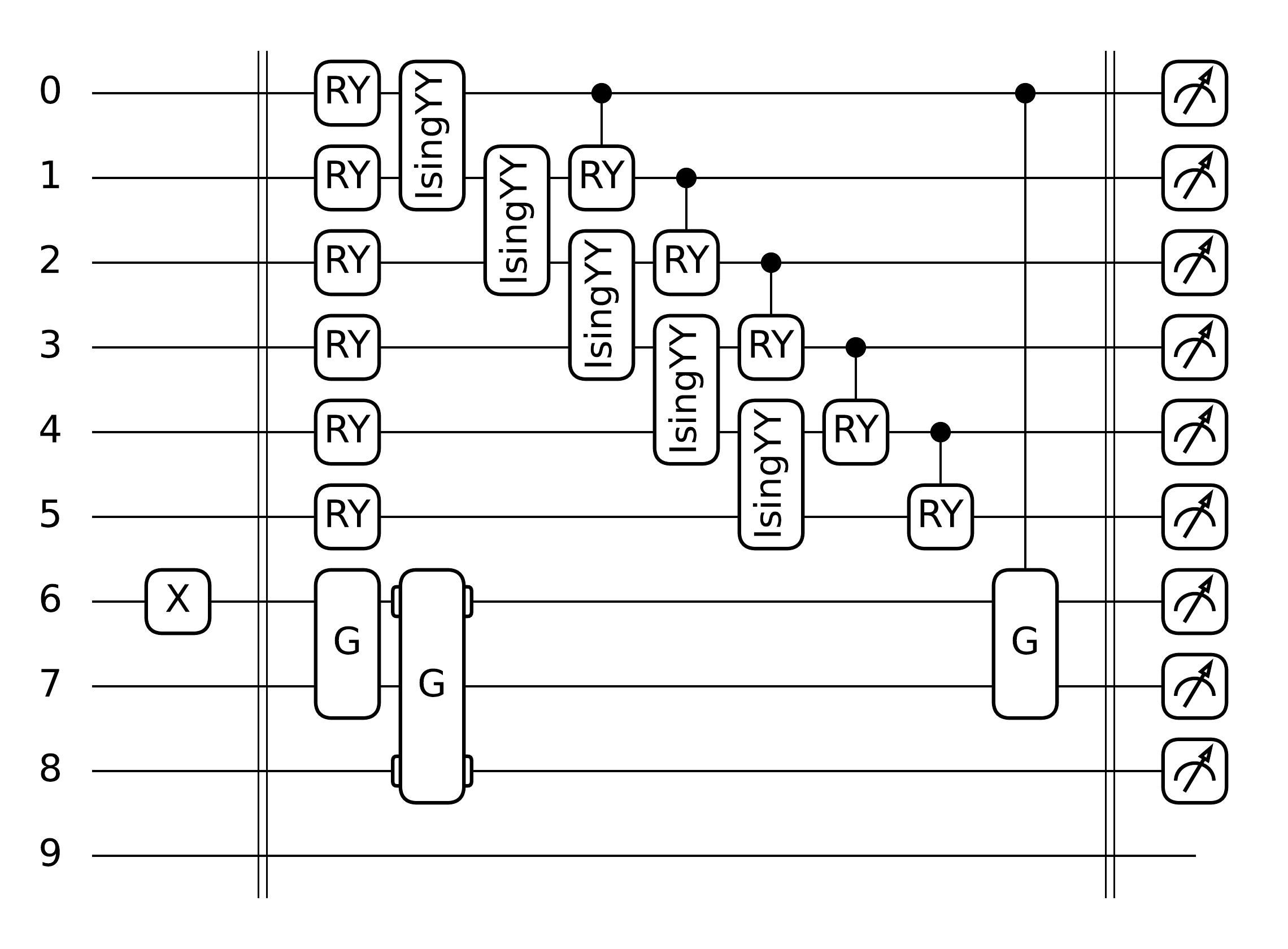}};
            %\node (fig1) at (4.5,0)
            %    {\includegraphics[height=4.85015 cm, clip, trim=35.1cm 0 0 0]{figures/bool.pdf}};
        
            % Insert \dots 10 times vertically aligned
            \foreach \i in {2,3,...,10} {
                \node at (3.15, -2.38+0.434*\i) {\tiny \dots};
            }
            % The layers
            \draw[decorate,decoration={brace,amplitude=4pt}] (-1.65, -2.29) -- (-2.45, -2.28) node[midway, below,yshift=-4pt,xshift=4pt]{\tiny initial excitation};
            \draw[decorate,decoration={brace,amplitude=4pt}] (2.76, -2.28) -- (-1.53, -2.29) node[midway, below,yshift=-4pt,]{\tiny model layer};
            % The register sizes
            \draw[decorate,decoration={brace,amplitude=4pt}] (-2.8, -0.34) -- (-2.8, 2.1) node[midway, below,xshift=-9pt, yshift=4.5pt]{\tiny n6};
            \draw[decorate,decoration={brace,amplitude=4pt}] (-2.8, -1.65) -- (-2.8, -0.46) node[midway, below,xshift=-9pt, yshift=4.5pt]{\tiny c3};
        \end{tikzpicture}
        \caption*{(b) Boolean design}
   \label{fig:bool} 
\end{minipage}

\caption{Non-Boolean circuit design (a) and Boolean circuit design (b) for an [n5, c3, c2] register. The model layer can be repeated $d$ times to obtain a depth-$d$ circuit. All qubits are measured in the computational basis to obtain a bitstring generated by the model. Here, the Boolean circuit design saves one qubit by treating the two-category feature as a Boolean variable and merging it into the numerical register.}
\label{fig:circuits} 
\end{figure}

Figure \ref{fig:circuits} shows the two variational circuit designs for the quantum generator: a \textit{non-Boolean} and a \textit{Boolean} design are proposed. In both designs, the circuit consists of a single $n$-qubit numerical register where all numerical features are binary-encoded, followed by multiple categorical registers, ordered by qubit count or number of categories they represent. In the non-Boolean design, all categorical features are one-hot encoded, whereas in the Boolean design, the encoding of binary categories can be optimized by replacing the one-hot encoding with a Boolean encoding and merging the Boolean variable into the numerical register. The Boolean design saves one qubit per two-category feature. For example, a register configuration in the non-Boolean circuit might be denoted as [n5, c3, c2], where n5 represents a numerical register with five qubits, c3 is a three-qubit one-hot register, and c2 is a two-qubit one-hot categorical register. In the Boolean circuit design, the same configuration could be simplified to [n6, c3], where the binary category is absorbed into the numerical register.

The upper part of the circuit represents an $n$-qubit numerical register that consists of a layer of RY rotations on each qubit, followed by pairwise IsingYY gates and controlled RY rotations \cite{pennylane}. The lower part of the circuit consists of multiple categorical registers. Each categorical register is initialized by an X-gate to prepare a reference state such as $\lvert1000\rangle$, followed by pairwise single-excitation gates. The entanglement between different registers is established by controlled single-excitation gates $G$ to capture correlations between different features.

\subsection{Training QGAN}\label{sec:training_qgan}

\begin{figure}[h!]
    \centering
    \includegraphics[width=0.9\linewidth]{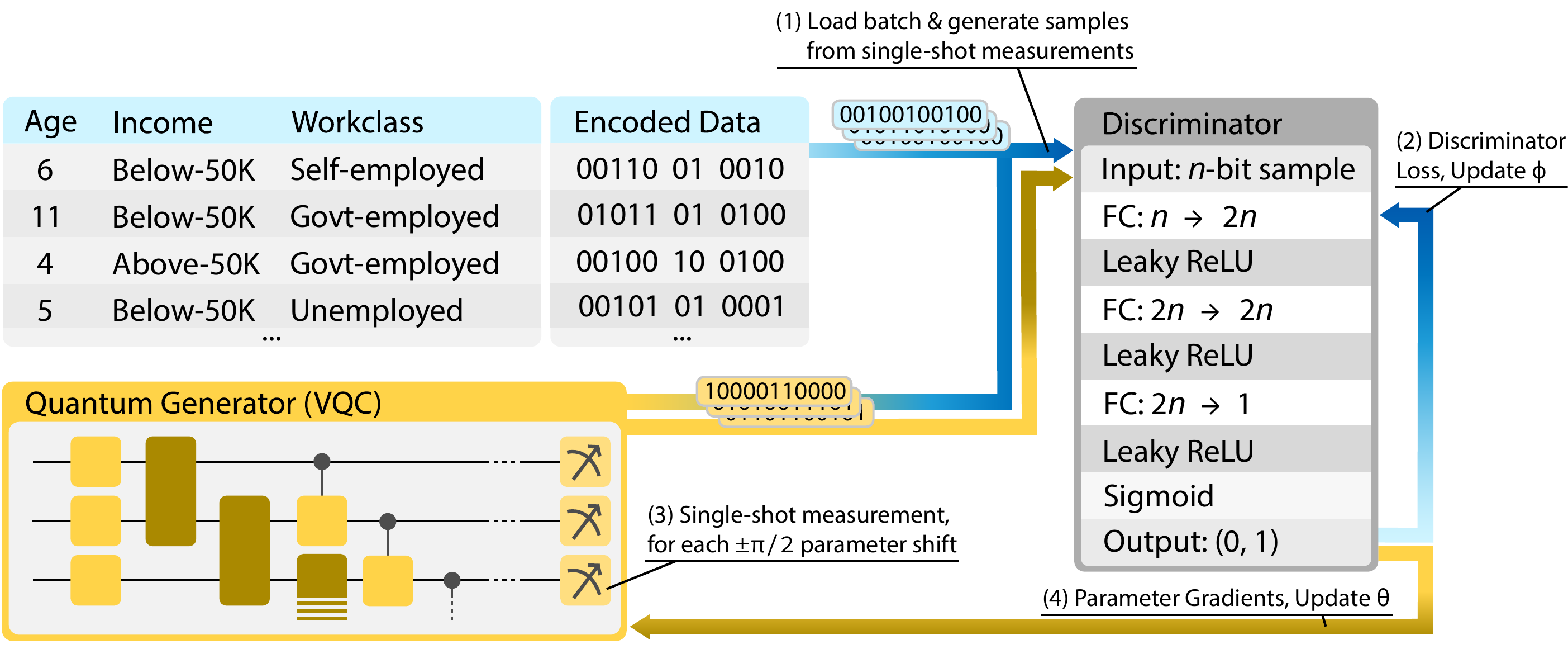}
    \caption{Schematic diagram of TabularQGAN training. In Step 1, either a batch of training data or a batch of synthetic samples (obtained from single-shot measurements) is fed to the discriminator. In Step 2, the discriminator attempts to distinguish between real and fake samples, and its parameters $\phi$ are updated based on the gradient of the discriminator loss~$L_{D}$. In Step 3, a sample is generated for each parameter shift, and the discriminator with fixed parameters $\phi$ is used to compute the gradient of the parameters according to the parameter-shift rule. In Step 4, the generator parameters~$\theta$ are updated based on their gradient.}
    \label{fig:Schematic-Diagram}
\end{figure}

Training of our TabularQGAN model proceeds by alternately updating the classical discriminator with a sigmoid output and the quantum generator. All quantum runs in this work were performed as exact statevector simulations on classical hardware. The training pipeline is described in Figure \ref{fig:Schematic-Diagram}.
First, the state is initialized as $|0 \rangle^{\otimes N}$ and the  VQC is initialized with random parameters. At each training iteration, a batch of $m$ (real) samples $x_i$ is drawn from the encoded training set, and the resulting state $\lvert\psi(\theta,z)\rangle$ is measured in the computational basis to yield bitstrings $x' = b_{N-1} b_{N-2} \dots b_0 $, which are then decoded into numerical and categorical values. These synthetic samples $x'$ are fed into a classical feed-forward network $D(x';\phi)$ that outputs $D(x')\in(0,1)$, which estimates the probability that the input is real. During each training iteration, we first update the parameter vector $\phi$ of the discriminator by minimizing the objective function \cite{goodfellow_generative_2014}
\begin{equation} 
\label{eq:loss_function}
L_D(\phi) = -\frac1m\sum_{i=1}^m\log D(x_i;\phi)\;-\;\frac1m\sum_{j=1}^m\log\bigl(1 - D(x'_j;\phi)\bigr)~,
\end{equation}
where $\{x_i\}$ are real records and $\{x'_j\}$ are generator outputs. Next, the updated VQC is executed and measured $m$ times to generate $m$ new synthetic samples $x^\prime_j$. The generator is updated by fixing $\phi$ and minimizing
\begin{equation}
L_G(\theta) = -\frac1m\sum_{j=1}^m\log D(x'_j(\theta);\phi)~,
\end{equation}
thereby encouraging $G(\theta)$ to produce samples that the discriminator labels as real. Gradients with respect to $\phi$ are computed via standard backpropagation. Gradients with respect to the quantum parameters $\theta$ are obtained using the parameter-shift rule \cite{quantumcircuitlearning, analytic_gradients_quantum_hardware}: for each parameter $\theta$, the gradient is evaluated as the derivative of the observable $\langle M\rangle$:
\begin{equation}
\label{eq:parameter_shift}
\frac{\partial \langle M \rangle}{\partial \theta} = \frac{1}{2} \left[ \langle M \rangle_{\theta + \frac{\pi}{2}} - \langle M \rangle_{\theta - \frac{\pi}{2}} \right]~,
\end{equation}
where $M = D(\text{measure}(\left|\psi(\theta, z)\right\rangle$)) \cite{QGANKilloran}. The training continues for $T$ epochs. A successfully converged model has a quantum generator that has learned to produce synthetic records indistinguishable from real tabular data by the classical discriminator. The training procedure is formally presented in Algorithm \ref{alg:qgan_training}.
\begin{algorithm}[h!]
\caption{TabularQGAN Training Algorithm}
\label{alg:qgan_training}

\textbf{Initialize:} Generator parameters $\theta$, Discriminator parameters $\phi$, batch size $m$, learning rates $\eta_G$, $\eta_D$, total epochs/training steps $T$, discriminator steps $k_D$

\begin{algorithmic}[1]
\For{$t = 1$ to $T$}
    \State \textbf{Discriminator Training:}
    \For{$i = 1$ to $k_D$}
        \State {Sample batch of $m$ examples:} $\{x_1,...,x_m\} \sim p_{\mathrm{real}}$
        \State Generate synthetic data:
            \State Prepare quantum state $\lvert \psi \rangle \leftarrow G(\theta)$
            \State Measure and decode: $x' \leftarrow \text{Measure}(\lvert \psi \rangle)$
            \State Compute discriminator loss:
            \[
            L_D(\phi) = -\frac{1}{m} \sum_{j=1}^m \log D(x_j; \phi) - \frac{1}{m} \sum_{k=1}^m \log(1 - D(x'_k; \phi))
            \]
            \State Update discriminator weights: $\phi \leftarrow \phi - \eta_D \nabla_\phi L_D(\phi)$
    \EndFor
    \State \textbf{Generator Training:}
        \State Generate synthetic data:
            \State Prepare quantum state for each ($\pm \pi/2$)-parameter shift $\lvert \psi \rangle \leftarrow G(\theta)$
            \State Measure and decode: $x'_j \leftarrow \text{Measure}(\lvert \psi \rangle)$
        \State Compute generator loss:
        \[
        L_G(\theta) = -\frac{1}{m} \sum_{j=1}^m \log D(x'_j; \phi)
        \]
        \State Update generator parameters with parameter-shift rule:
        \[
        \theta \leftarrow \theta - \eta_G \nabla_\theta L_G(\theta)
        \]
\EndFor
\end{algorithmic}
\end{algorithm}

% Mention that we have used a SIMULATOR and not a Quantum Computer to obtain the results! Here, or in the hyperparameter optimization section, I'd mention which simulator we have used (which qml.device(...)).

% discriminator: optax.adam @ learning_rate 0.1, b1 = 0.5, b2 = 0.999
% Discriminator_JAX(), weights initialized randomly, dummy (random) input fed into discriminator during initialization

% generator: optax.sgd @ learning_rate 0.1, 
% generator weights initialized with jax.uniform * pi; shift every parameter with ±1

\subsection{Qubit and Gate Complexity} \label{sec:resource_analysis}

Consider a variational block applied to a configuration \([\;\text{n}x_0 ,\,\text{c} x_1,\,\text{c} x_2,\dots,\text{c} x_{R-1}\;]\) with \(R\) registers (1 numerical + \(R-1\) categorical). The number of gates required for a numerical register (\(x_0\) qubits) is given by 
\begin{equation}
      g_{\mathrm{num}}(x_0)
      = \underbrace{x_0}_{R_y}
      +\underbrace{(x_0-1)}_{\text{Ising-YY}}
      +\underbrace{(x_0-1)}_{\text{c-}R_y}
      = 3x_0 - 2~.
\end{equation}
  
For each categorical register of size \(x_i\), the number of gates is equal to the number of qubits or the size of the register
\begin{equation}
      g_{\mathrm{cat}}(x_i)
      = \underbrace{1}_{X\text{-prep}}
      + \underbrace{(x_i - 1)}_{\substack{\text{single-}\\\text{excitation}}}
      = x_i~.
\end{equation}

Cross-register entanglers are controlled single-excitation gates between each adjacent pair of registers
\begin{equation}
      g_{\mathrm{cross}}(R) = R - 1~.
    \end{equation}
    Hence, the total gate count is
    \begin{equation}
  g_{\mathrm{total}}
  = g_{\mathrm{num}}(x_0)
  + \sum_{i=1}^{R-1} g_{\mathrm{cat}}(x_i)
  + g_{\mathrm{cross}}(R)
  = \bigl(3x_0-2\bigr) + \sum_{i=1}^{R-1} x_i + (R-1)~.
\end{equation}

\textbf{Example:} [n5,c3,c2], (\(R=3\)) \[g_{\mathrm{num}}(5)=3\cdot5-2=13~,\quad g_{\mathrm{cat}}(3)=3~,\quad g_{\mathrm{cat}}(2)=2~,\quad g_{\mathrm{cross}}(3)=2~,\quad\] \[ g_{\mathrm{total}}=13+3+2+2=20~.\]

Let \(x_0\) be the number of numerical qubits, \(m=\sum_{i=1}^{R-1}x_i\) be the total number of categorical qubits, $R$ be the total number of registers, and \(N=x_0+m\) be the overall qubit count.  Since \(R\le N\), we have
\begin{equation}
  g_{\mathrm{total}}
  = 3x_0 - 2 + m + (R-1)
  \;=\; \mathcal{O}(x_0 + m + R)
  \;=\; \mathcal{O}(N)~.
\end{equation}
Thus, the total gate count scales \emph{linearly} with the total number of qubits \(N\). For a circuit with $d$  repeated layers (depth $d$), the total gate count is  $\mathcal{O}(d \cdot N)$. 

\subsection{Training Cost Analysis}
The analysis above establishes the gate complexity
for a depth-$d$ circuit. However, this does not fully capture the training cost, and it is important 
to distinguish between two execution contexts: classical simulation as used in this work, and future 
execution on quantum hardware. Here, we define cost as the number of operations per execution setting: circuit evaluations in the classical simulation setting and gate operations in future quantum hardware execution.

Since the quantum generator is trained using the parameter-shift rule (Eq.~\ref{eq:parameter_shift}), 
each gradient update requires two circuit evaluations per trainable parameter, one evaluated at 
$\theta + \frac{\pi}{2}$ and the other at $\theta - \frac{\pi}{2}$. For a circuit with 
$P = \mathcal{O}(d \cdot N)$ trainable parameters, this gives $2P = 2 \cdot \mathcal{O}(d \cdot N)$ 
circuit evaluations per generator gradient step, i.e.,

\begin{equation}
    \text{Circuit evaluations per generator step} = 2P \approx \mathcal{O}(d \cdot N)~.
    \label{eq:circuit_evals}
\end{equation}

In the present work, all circuits are executed as noiseless statevector simulations 
on classical hardware using the PennyLane library~\cite{pennylane}. On a classical 
statevector simulator, the quantum state must be stored and manipulated as a vector of $2^{N}$ 
complex amplitudes. Applying a single gate therefore requires updating $\mathcal{O}(2^{N})$ 
amplitudes, resulting in $\mathcal{O}(d \cdot N \cdot 2^{N})$ operations per full circuit evaluation. The total simulation cost per training step is therefore:

\begin{equation}
\begin{aligned}
\text{Total cost per training step}
&= \text{Circuit evaluations per step} \times \text{Cost per evaluation} \\
&= \mathcal{O}(d \cdot N) \cdot \mathcal{O}(d \cdot N \cdot 2^{N}) \\
&= \mathcal{O}(d^{2} \cdot N^{2} \cdot 2^{N})~,
\end{aligned}
\label{eq:training_cost}
\end{equation}
which scales exponentially with the number of qubits $N$. This exponential overhead is inherent to classical simulation of quantum circuits and is precisely why the present study is restricted to 10-15 qubits. We do not claim any computational advantage over classical models in this simulation-based setting; rather, the experiments serve as a proof-of-concept to validate the model's ability to learn the correct data distributions.

On actual quantum hardware, the classical statevector overhead does not apply, and each circuit evaluation costs $\mathcal{O}(d \cdot N)$ gate operations. In this hardware execution regime, the $2P = 2 \cdot \mathcal{O}(d \cdot N)$ gradient evaluations per training step may represent a substantially smaller training cost compared to classical GAN baselines, because purely classical models typically require parameter counts that are orders of magnitude larger (e.g., $131{,}072$ parameters for CTGAN versus $88$ for TabularQGAN on the MIMIC~10 dataset, as shown in Table~\ref{tab:overall_result}). The total gate cost per training step on quantum hardware would reduce to

\begin{equation}
    \text{Total gate cost per step (hardware)} = 2 \cdot \mathcal{O}(d \cdot N) \cdot 
    \mathcal{O}(d \cdot N) = \mathcal{O}(d^{2} \cdot N^{2})~,
    \label{eq:hardware_cost}
\end{equation}
which scales only polynomially with $N$. Whether this parameter efficiency translates to a practical training advantage on near-term or fault-tolerant quantum hardware remains an open question and is an important direction for future work.

\subsection{Evaluation and Benchmarking} \label{sec:eval_benchmarking}

\subsubsection{Benchmarks} We benchmark our quantum generator against several classical baselines: CTGAN and CopulaGAN are two adaptations of the well-known GAN architecture \cite{NIPS2014_5ca3e9b1} with additional data preprocessing techniques. CTGAN is introduced in Reference~\cite{xu2019modeling} and CopulaGAN via the SDV library \cite{sdv}. Both models are trained by minimizing the loss function in Eq. \ref{eq:loss_function}. We also benchmark against two additional classical architectures, both popular for generative tasks: a large language model (LLM) and a variational autoencoder (VAE). be-GReaT is a large-language-model-based architecture introduced in \cite{be-great}, which we use with OpenAI's GPT-o4-mini model \cite{openai_o4mini}. 
VAE-GMM\cite{apellaniz2024improved} is a variational autoencoder that uses a (Bayesian) Gaussian Mixture Model (GMM) to give a more complex representation of the latent space.
As both of these models can natively ingest data that is a mixture of strings, floats, and integers, the training data was not preprocessed into (or evaluated in) binary form but was instead left in human-readable format, e.g., for Adult Census 10, a row might read `16, Above-50K, govt-employed'; see \ref{sec:preprocessing} for further details.

\subsubsection{Evaluation Metrics} \label{sec:metrics} We evaluate the performance of the models using three complementary measures: an overall similarity score from SDMetrics  \cite{sdmetrics}, the overlap fraction between the training data and synthetic samples, and the final metric measures downstream predictive performance on generated data.   
The SDMetrics overall similarity score~\cite{sdmetrics} is an average over two components. The first is a column-wise measure of univariate marginals, the column shape similarity (see Appendix \ref{sec:sd_definition_score}).
The second component is column pair trends, which capture bivariate relations. 
In addition to these statistical similarity metrics, we also evaluated two measures of the models' generalization performance.
The first is the overlap fraction between the training data and the synthetic data, defined as $1~-~U_{S \notin R}/U_S $ where $U_{S \notin R}$ is the count of unique rows in the synthetic data that are not present in the training data and $U_S$ is the number of unique samples in the synthetic data. Hence, an overlap fraction of one would imply that all samples in the synthetic data were also found in the training data. This metric can be interpreted as an indication of memorization: high overlap suggests stronger retention of exact training records, while lower overlap suggests that the generator produces more samples not exactly present in the training set. However, the overlap fraction is only a coarse heuristic, and a low value does not constitute a privacy guarantee, since training records could still reveal sensitive information through attacks such as membership inference \cite{ShokriSS16}. A high overlap, however, is a definite concern, because the generated samples cannot be considered distinct from the training data. This limitation is particularly relevant in healthcare settings, where rare combinations of demographic, diagnostic, or admission-related attributes may be identified even when no generated record is an exact copy of a training record. We note that in this study, we restrict ourselves to considering the overlap metric and do not perform a membership inference attack or similar formal privacy evaluation on the models.

The second generalization metric, which we call the downstream score, measures how well synthetic data can effectively replace real data in a classical supervised learning task. For each dataset and feature-target combination, we train an XGBoost model \cite{xgboost} on both real and synthetic data using identical hyperparameters and training procedures. XGBoost is an optimized distributed gradient boosting library based on tree ensemble algorithms. Unlike traditional ensemble methods such as Random Forests that train deep decision trees independently in parallel, XGBoost builds shallow decision trees sequentially. Each subsequent tree is trained to predict the residuals (errors) of the preceding trees, systematically minimizing a regularized objective function using a gradient descent optimization framework. XGBoost was selected as the benchmark classifier and regressor due to its state-of-the-art performance on tabular data, robustness to varying data scales, and widespread adoption in production machine learning workflows. We used the library XGBoost\cite{xgboost} for implementation and employed a hyperparameter optimization procedure via a grid search. This grid search is executed independently for each real and synthetic training set, tuning tree-structure parameters (such as learning rate, maximum tree depth, and training-instance subsampling ratio) to ensure each model achieves optimal convergence. If the target is categorical, we report the classification accuracy, and for numerical columns, we report the coefficient of determination $R^2$. The downstream score is then defined as the absolute difference between these two performance values: $ \text{Downstream Score} \;=\; \bigl|\text{Score}_{\mathrm{real}} \;-\; \text{Score}_{\mathrm{synthetic}}\bigr|$. A near-zero score indicates that the synthetic data faithfully preserves the predictive relationships found in the original dataset. 
These metrics collectively quantify the quality and usability of synthetic samples across statistical and downstream learning dimensions.
These evaluation metrics were all calculated from $N$ samples (where $N$ is the length of the original training data) from the trained model. For the TabularQGAN, the samples were inverse-transformed back into human-readable text, and for the classical models, this was not required, as their direct outputs had this form. All metric calculations on each model were therefore performed on the same data representation.

\section{Experimental results} \label{sec:experimental_results}

% Experimental Results (2 pages)
% 	1. Experimental setup ( Model configuration, Hyperparameters, experimental runs)
%   2. Experimental Results
%        Results of the quantum model with different datasets
%        Results of classical models with the same datasets
%   3. Analysis of quantum model against classical problems like Mode collapse, etc(Also, barren plateaus if seen in our quantum model training)
%
%
%
%
%
%
\subsection{Data sets} \label{sec:datasets}
We evaluate our model on two standard datasets, the MIMIC-III clinical dataset \cite{johnson_mimic-iii_2016} and the Adult Census income dataset \cite{adult_census}. MIMIC-III is a publicly available anonymized health-related data set with 100,000 samples of over forty thousand patients, and the Adult Census contains income records of adults including information on age education and work type, with 35,000 samples. Due to the high computational resources required for simulating quantum computers, we take subsets of each dataset's features such that each dataset contains numerical and categorical features and is divided into 10 and 15-qubit datasets. The number of qubits and features used for each dataset configuration is shown in Table \ref{tab:dataset_overview}.

\subsection{Hyperparameter Optimization}\label{sec:hyper_param_opt}
Hyperparameter optimization was performed over all four data sets for both the quantum and classical models. The optimization was performed via a grid search over the hyperparameters. For the classical and quantum GAN models, these were circuit depth, batch size, generator learning rate, discriminator learning rate, and, for the classical models only, layer width. The layer width was either a fixed value of 256 or data set dependent, as twice the dimension of the training data. Some of the hyperparameter ranges differ between the classical models and the TabularQGAN. This was due to initial experiments indicating which ranges led to better performance. Each model configuration was repeated five times with a different random seed. Due to the different structures of the classical VAE and LLM models, different hyperparameters were varied, with the goal of having the total number of runs approximately equal. The values of the parameters varied are shown in Table \ref{tab:hyperparameter_grid}. For each model, the best hyperparameter settings were selected with respect to the overall metric defined in section \ref{sec:metrics} by aggregating over 5 seeds of each hyperparameter setting and taking the mean, the resulting configurations can be found in Appendix \ref{sec:opt_hyperparameter}. To ensure a fair comparison, training durations were adapted to the convergence behaviour of each model. The quantum model was trained for a maximum of 3000 iterations, translating to either 300 or 600 epochs depending on the batch size, with the optimal performance selected via checkpoint analysis. The SDV implementation of CopulaGAN and CTGAN lacks automated checkpointing. Instead, we performed a systematic grid search across 1500 epochs at 10-epoch intervals to identify the peak performance window and choose this for the benchmarking. Similarly, the VAE-GMM model was trained for 200 epochs, after which stable convergence was observed. For the LLM model, stable convergence is observed after a single epoch comprising 4000 iterations. The much lower epoch count can be explained by the fact that the model was pretrained on tabular data and, unlike the other models, was not trained from scratch. No other early stopping criteria were used. The number of hyperparameter configurations per model type and data set pair is equal (360 configurations per model), with the exception of the CTGAN and CopulaGAN configurations, which had an additional setting of `Layer Width', meaning that they had 720 configurations each. 

{\scriptsize
\begin{longtable}{>{\raggedright\arraybackslash}m{2.7cm}>{\centering\arraybackslash}m{2.7cm}>{\centering\arraybackslash}m{3.2cm}>{\raggedright\arraybackslash}m{3.7cm}}

\caption{Overview of data sets used in experiments, including the number of qubits, numerical and categorical features.}
\label{tab:dataset_overview}
 \\ \hline
\textbf{data set Name} & \textbf{Number of Qubits} & \textbf{Numeric Features} & \textbf{Categorical Features}\\ \hline
\endfirsthead
%rows
    Adult Census 10 & 10 & Age & Income, Education \\ 
    Adult Census 15 & 15 & Age & Workclass, Education \\ 
    Mimic 10 & 10 & Age & Gender, Admission type\\ 
    Mimic 15 & 15 & Age, Admission time & Gender, Admission type \\ \hline
\end{longtable}
}

\subsection{ Overall Metric Results} \label{sec:numerical_results}

\begin{figure}[h]
\centering
\begin{subfigure}{0.47\linewidth}
    \centering
    \includegraphics[trim={40pt 0pt 40pt 0pt},clip,width=\linewidth]{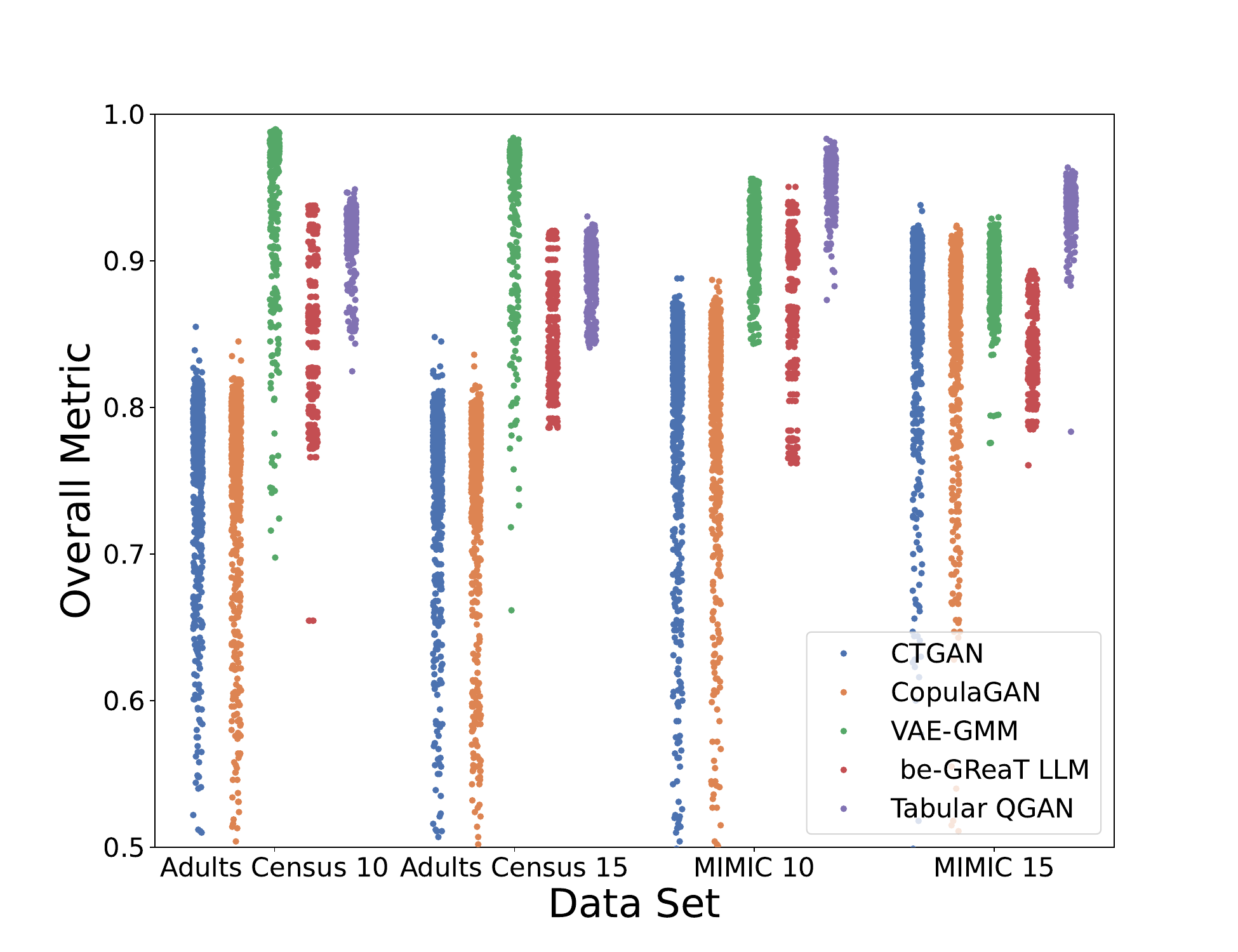}
    \caption{Plot of the overall metric for each hyperparameter configuration and seed for each data set. The spread of the points within each bar is artificially added to improve data visibility.}
    \label{fig:main_result_overall_metric} 
\end{subfigure}
\hfill
\begin{subfigure}{0.52\linewidth}
    \centering
    \includegraphics[trim={0pt 0pt 0pt 0pt},clip,width=\linewidth, height=.77\linewidth]{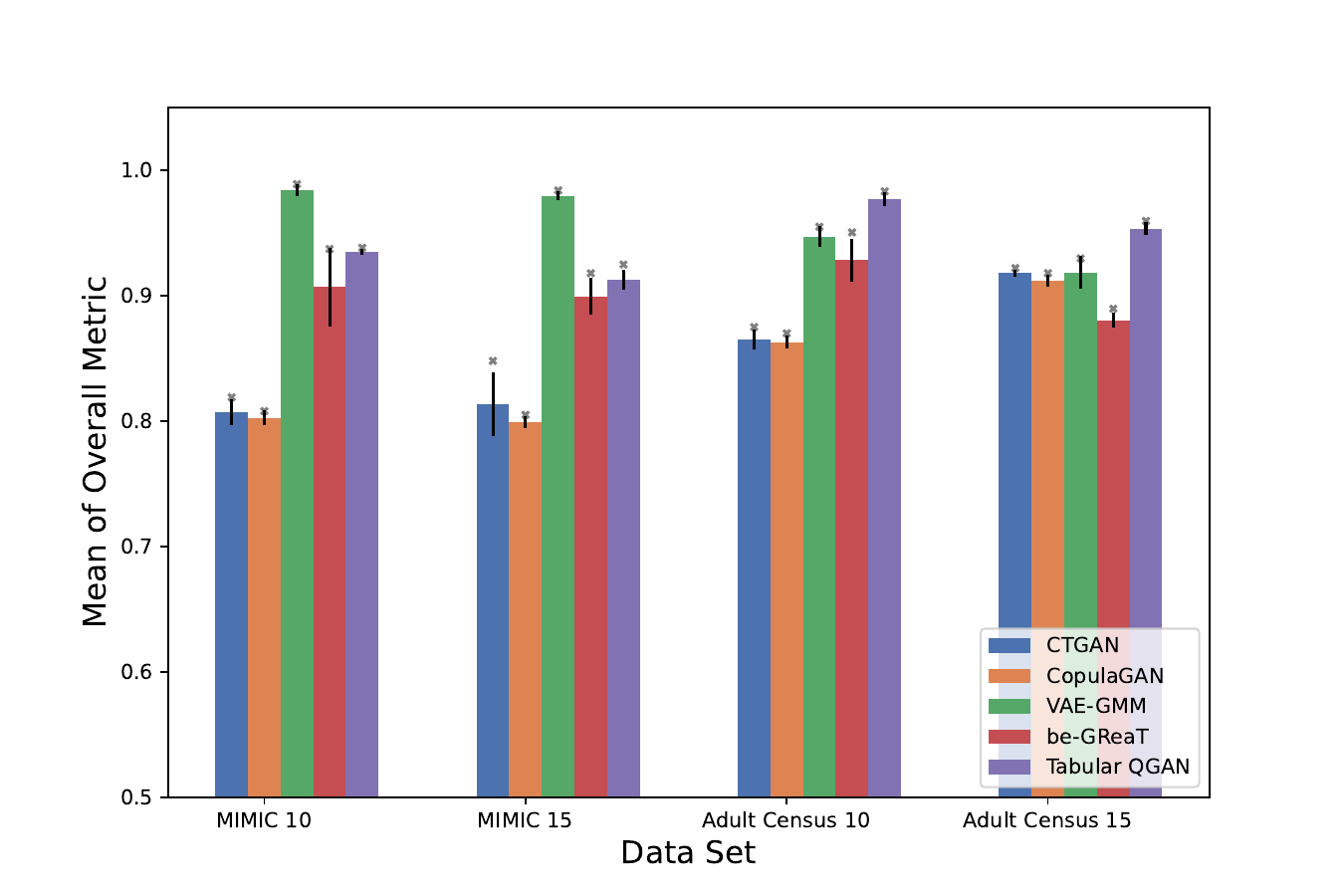}
    \caption{Bar chart of the overall metric for each best found hyperparameter configuration averaged over seeds, the error bar is the standard deviation, and the grey cross is the maximum over seeds found.}
    \label{fig:best_config_comparsion}
\end{subfigure}
\caption{ } 

\end{figure}

In this Section, we discuss the results of the quantum and classical models with respect to the metrics introduced in Section \ref{sec:eval_benchmarking}. Our experiments are conducted using BASF's HPC cluster Quriosity, on CPU nodes. All quantum models are executed on noiseless state vector simulations using the PennyLane library~\cite{pennylane}. Additional key machine-learning libraries used were PyTorch and JAX. Although we described two circuit topologies in Section \ref{sec:Quantum_Generator}, our experiments across four data sets indicate that Boolean and non-Boolean encodings yield comparable performance, see Appendix \ref{sec:circuit_encoding} for more details. Therefore, in this section, we focus exclusively on the results obtained from Boolean-encoding models.
The first column in Table \ref{tab:overall_result} shows the results of the overall similarity metric for different models, averaged over seeds. Figure~\ref{fig:best_config_comparsion} shows this in bar-chart form. A score of 1 indicates perfect similarity between the probability distributions of the synthetic and training data sets, whereas 0 implies no similarity. 
The TabularQGAN model performs best overall for the MIMIC-III data, and VAE-GMM for the Adult Census data, see Figure \ref{fig:main_result_overall_metric}. The LLM model performs similarly to TabularQGAN for the Adult Census data and worse for the MIMIC-III data. We note that our  TabularQGAN outperforms the classical GAN-based models, CTGAN and CopulaGAN, which are the closest classical counterparts. These results suggest that the proposed quantum neural network architecture and tabular-specific ansatz may contribute to the observed performance improvements. However, because the compared methods differ in several aspects, including model architecture, data representation, and sampling mechanisms, the present experiments do not isolate the specific source of these differences. We cannot attribute the observed performance gains to any single component of the proposed approach. This observation is in line with the idea from Section \ref{sec:methodology}  that the enhanced expressivity of the quantum circuit and the constrained search space induced by Givens rotations may contribute to its improved performance. However, the black-box nature of GAN models makes it challenging to directly verify this, and further ablation studies are required to directly assess the contribution of these factors. We can make an overall comparison of performance over data sets by taking the average of the percentage difference between the best classical benchmark score and the TabularQGAN for the overall metric score. We find that the VAE-GMM outperforms by an average of 4.4\%. 

% \begin{figure}[h!]
%     \centering
%     \includegraphics[width=0.7\linewidth]{figures/overall_metric_comparasion_all.pdf}
%     \caption{Plot of the overall metric for each hyperparameter configuration for each data set. The spread of the points within each bar is artificially added to improve data visibility.}
%     \label{fig:main_result_overall_metric}
% \end{figure}

\subsection{Overlap Fraction Metric and Downstream Score Results}

\begin{figure}[h]
\centering
\begin{subfigure}{0.49\linewidth}
    \centering
    \includegraphics[trim={40pt 0pt 40pt 0pt},clip,width=\linewidth]{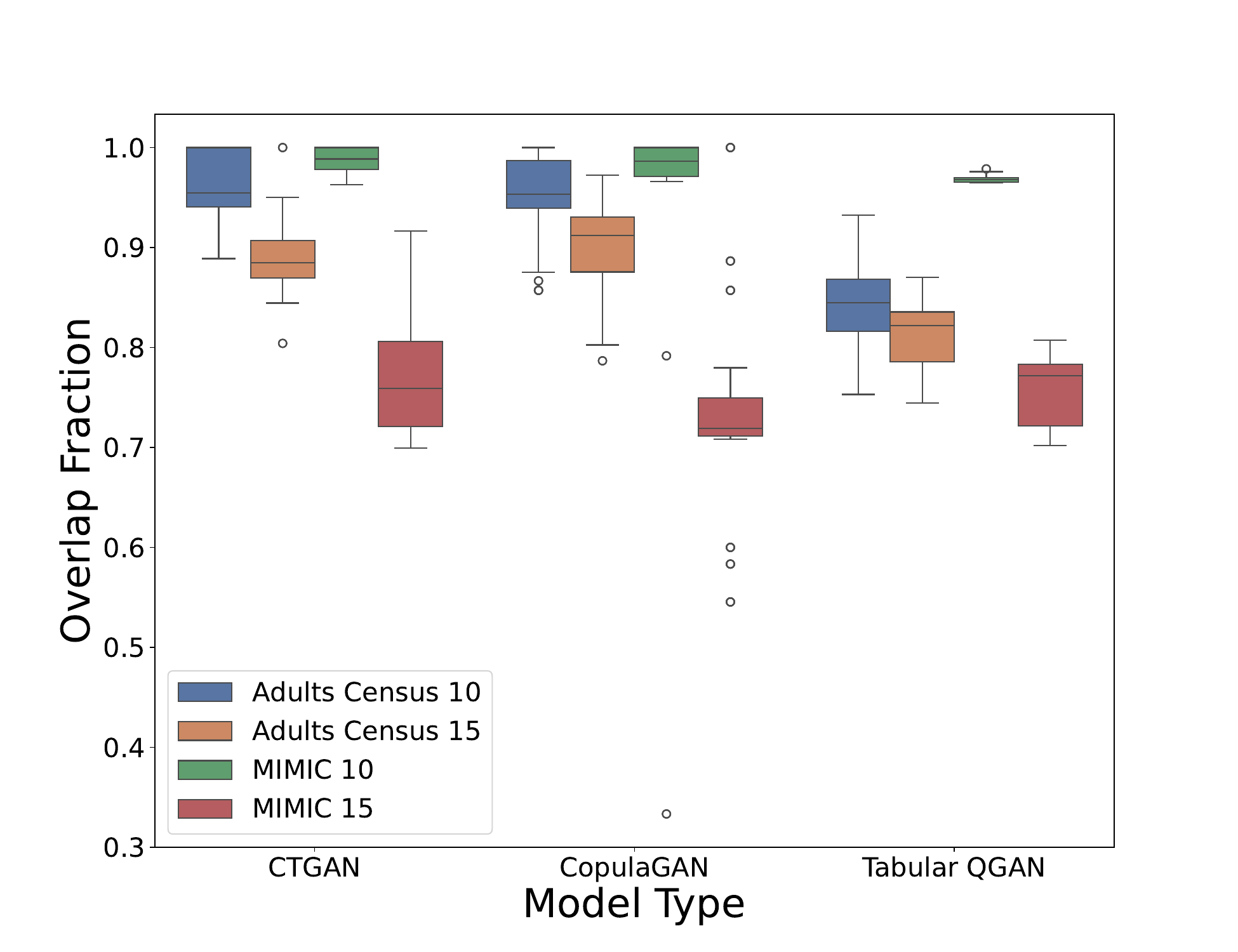}
    \caption{}
    \label{fig:overlap_gan} 
\end{subfigure}
\hfill
\begin{subfigure}{0.49\linewidth}
    \centering
    \includegraphics[trim={40pt 0pt 40pt 0pt},clip,width=\linewidth]{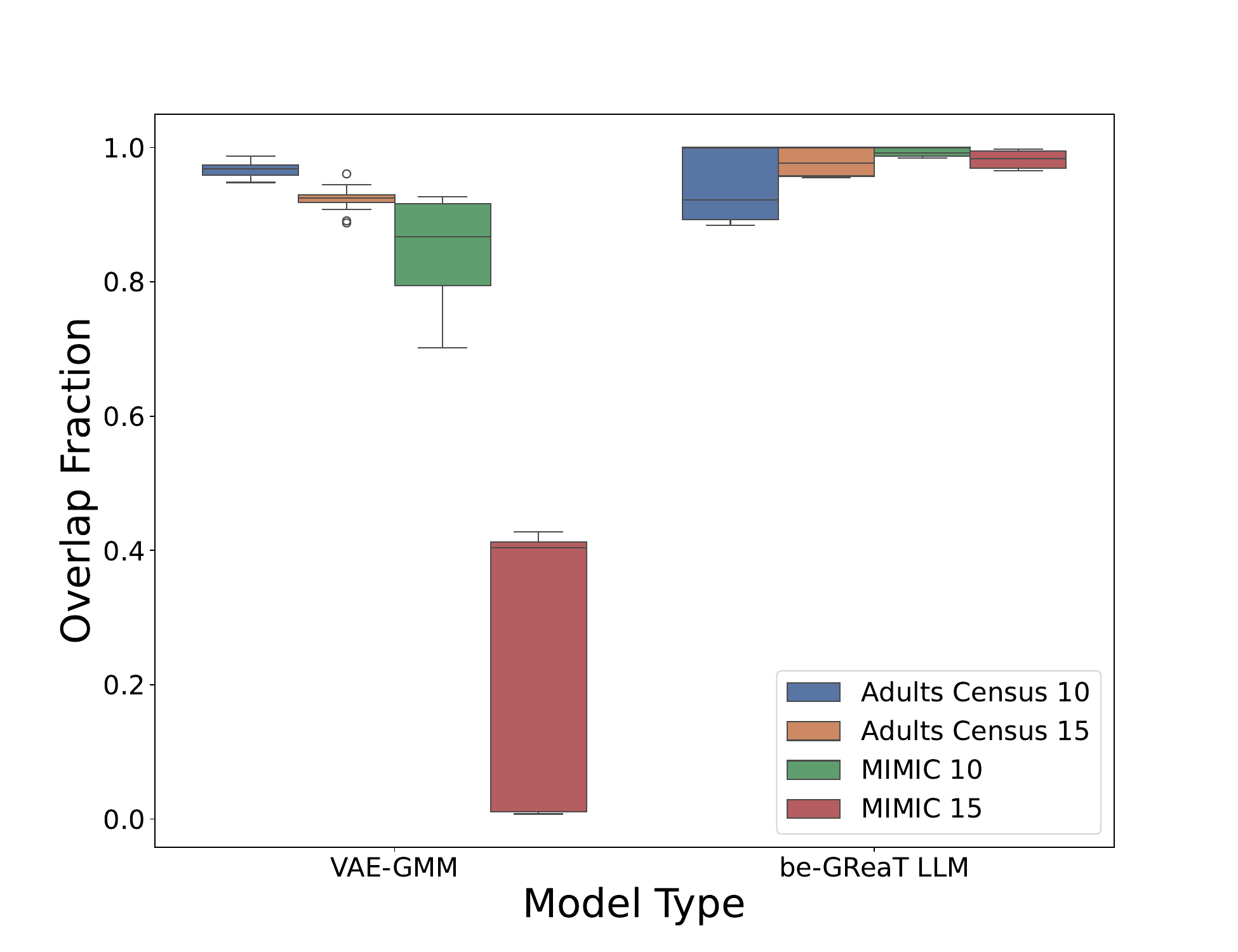}
    \caption{}
    \label{fig:overlap_vae_llm}
\end{subfigure}
\caption{Plot showing the overlap fraction metric for (a) GAN-type models and (b) VAE and LLM models. As explained in the text, only a selected subset of the data is sampled.}
\label{fig:overlap_fraction}
\end{figure}

Figure \ref{fig:overlap_fraction} is a bar plot of the overlap fraction for different model types. It shows that the overlap fraction is well over 50\% and close to 1 for some models and data sets. However, we find that the TabularQGAN model has on average a lower overlap score compared to classical models for the Adult Census data set but higher for the MIMIC data set. We find that the overall metric score and the overlap fraction are not correlated for either the classical or quantum models for any data set. Additionally, we find that for the LLM model, many of the overlap scores are exactly 1.0, meaning that the model perfectly reproduces the training data and that all generated samples are included in the training data set. The lower overlap score for the MIMIC 15 data set for the VAE-GMM is due to the presence of the `Admission Time' feature, which for the VAE-GMM data is encoded as a float and hence has a much wider range of possible values, reducing the overlap fraction. Due to the nature of the metric, interpreting a model with a high value as better than a model with a low value is not necessarily correct, as it is dependent on the intended use of the synthetic data, for example, for privacy preservation versus more similarity in the correlations between features. Hence, we include this metric to highlight the difference between the models in this aspect and to allow practitioners to decide how useful TabularQGAN is for their use case. 

\begin{figure}[h!]
\begin{subfigure}[h]{0.5\linewidth}
   \includegraphics[width=1.1\linewidth]{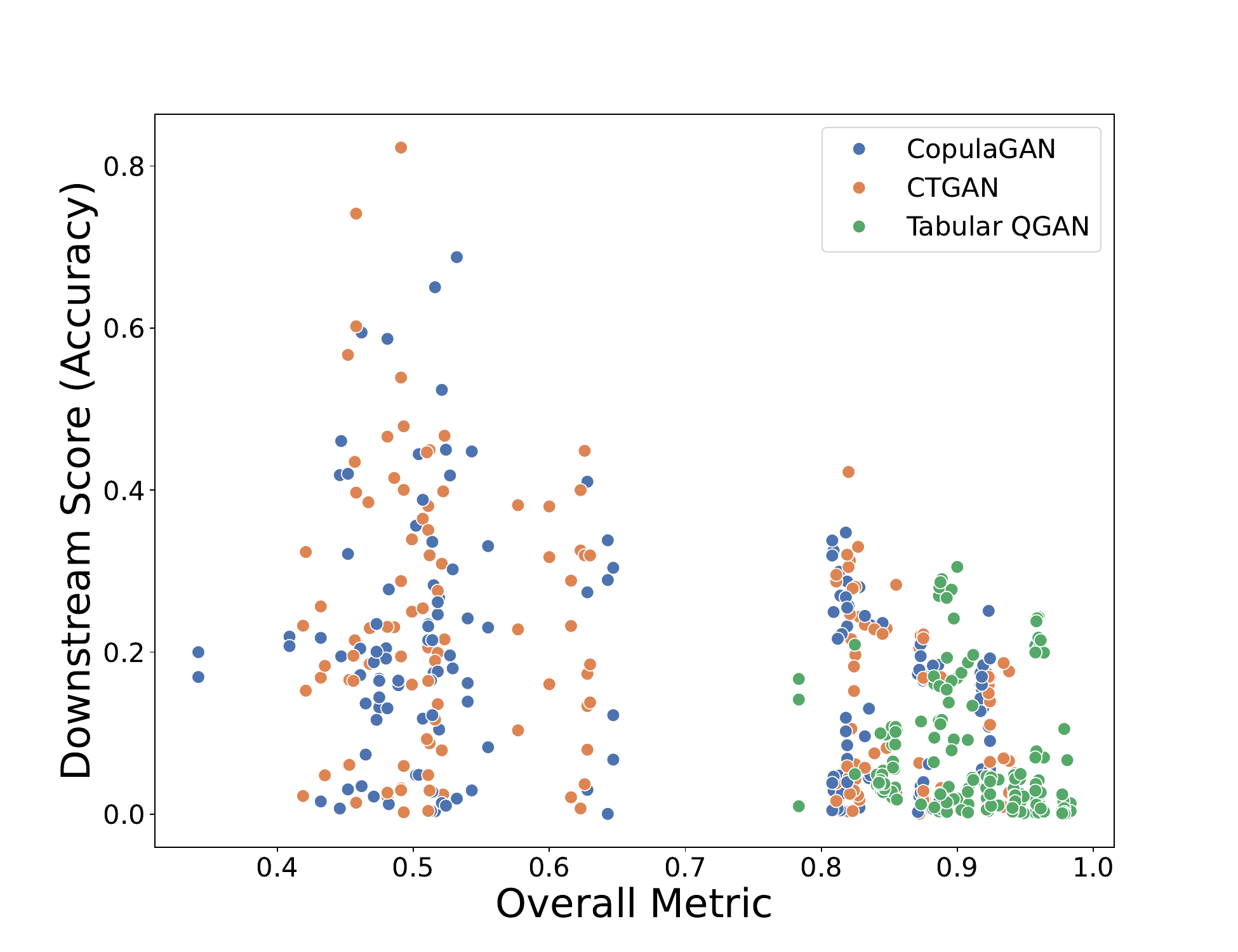}
   \caption{}
   \label{fig:downstream_score_cata} 
\end{subfigure}
\hfill
\begin{subfigure}[h]{0.5\linewidth}
   \includegraphics[width=1.1\linewidth]{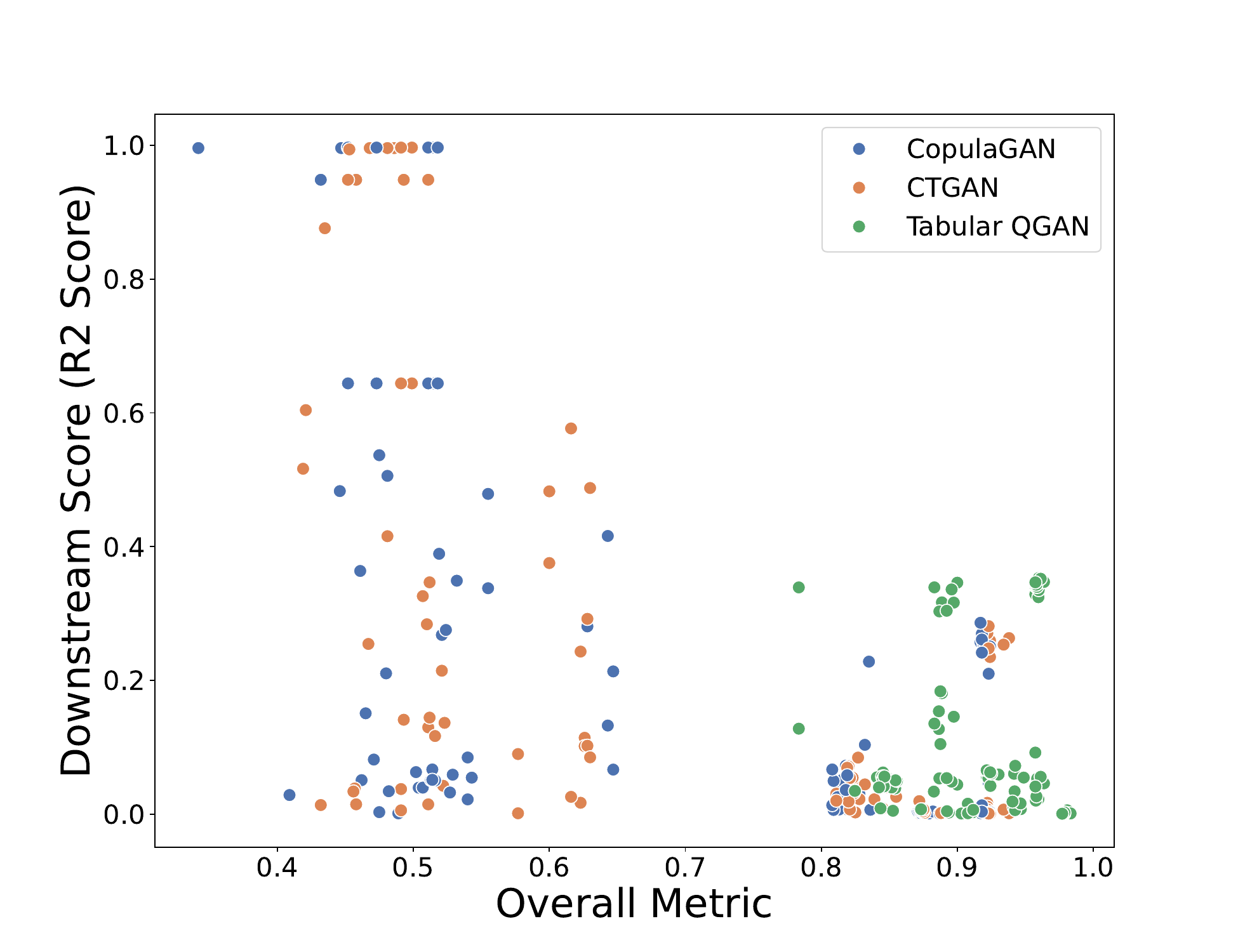}
   \caption{}
   \label{fig:downstream_score_numeric}
\end{subfigure}

\begin{subfigure}[h]{0.5\linewidth}
   \includegraphics[width=1.1\linewidth]{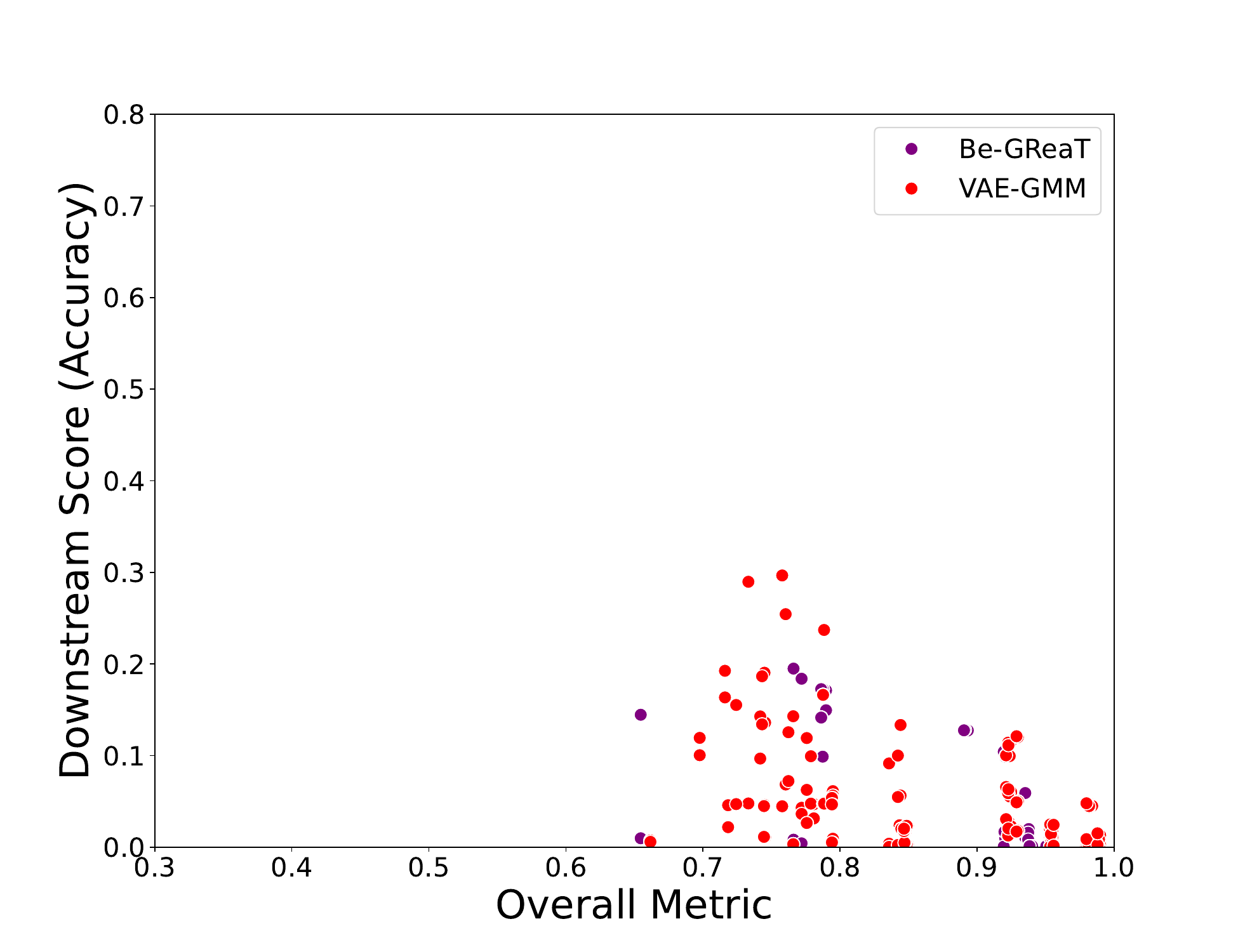}
   \caption{}
   \label{fig:downstream_score_cata_vae_llm}
\end{subfigure}
\hfill
\begin{subfigure}[h]{0.5\linewidth}
   \includegraphics[width=1.1\linewidth]{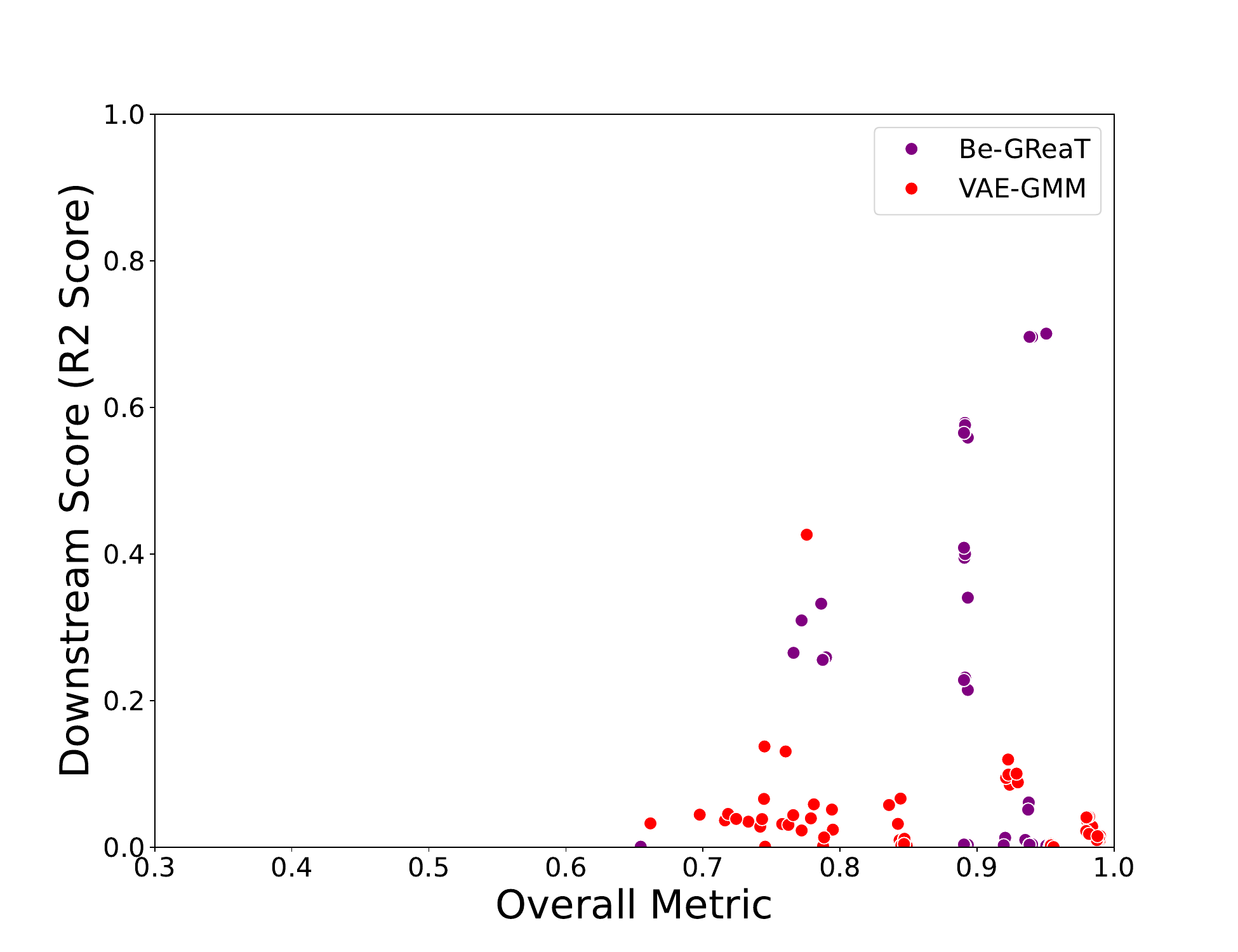}
   \caption{}
   \label{fig:downstream_score_numericvae_llm}
\end{subfigure}
\caption{ Downstream Score against Overall Metric for predicting categorical Variables for GANs (a) and VAE and LLM models (c). 
 Downstream Score against Overall Metric for predicting numerical Variables. for GANs (b)  and VAE and LLM model (d).   Only a selected subset of the data is sampled. }
 \label{fig:downstream}
\end{figure}

We plot the downstream metric against the overall metric in Figure \ref{fig:downstream}. We split the results into those for classification, Figure \ref{fig:downstream_score_cata} and Figure \ref{fig:downstream_score_cata_vae_llm} (where the target features are categorical), and regression tasks, Figure \ref{fig:downstream_score_numeric} and Figure \ref{fig:downstream_score_numericvae_llm} (where the target features are numerical). Again, these results are for a subsample of the overall hyperparameter-optimized data. We find that, on average, the TabularQGAN, VAE, and LLM models had a lower downstream score, indicating that those samples were better able to replicate the real data for training a classifier. This generalization metric is also contextually useful, for example, in a scenario where the original data, such as electronic health records, cannot be shared due to privacy concerns, so that a synthetic data set can be used instead. 

\begin{table}[h!]
\centering
\caption{Best performing models with respect to the overall metric for each data set and model type. The errors on the overall metric are the standard deviation over the seeds for that configuration. The number of parameters is stated for the best-performing hyperparameter configuration. The parameter count of OpenAI's GPT-o4-mini, the base model of be-GReaT, is not disclosed, but we report a typical range for modern LLMs \cite{brown2020languagemodelsfewshotlearners}.}

\vspace{0.6em} 
\label{tab:overall_result}
{\scriptsize
\begin{tabular}{llllll}
\hline
\textbf{Data Set}    & \textbf{Model Name} & \begin{tabular}[c]{@{}l@{}}\textbf{Overall} \\ \textbf{Metric}\end{tabular} & \begin{tabular}[c]{@{}l@{}}\textbf{Overlap} \\ \textbf{Fraction}\end{tabular} & \begin{tabular}[c]{@{}l@{}}\textbf{Downstream} \\ \textbf{Score}\end{tabular} & \begin{tabular}[c]{@{}l@{}}\textbf{Number of} \\ \textbf{Parameters}\end{tabular} \\ \hline 

\multirow{4}{*}{Adult Census 10} 
& TabularQGAN                    & $0.935 \pm     0.002      $           & 0.869                                 & 0.026            & 80  \\ 

& CTGAN                           & $ 0.807 \pm 0.010$                             & 0.953                                 & 0.112            & 131,072 \\ 
& CopulaGAN                       & $0.803 \pm 0.006$                               & 0.953                                 & 0.105            &  65,536   \\
& VAE-GMM                       & \bm{$0.984 \pm 0.005$}                               & 0.956                                 & 0.005           &  1,402   \\
& be-GReaT LLM                        & $0.907 \pm 0.031$                               & 0.903                                 & 0.025
           &  $10^9-10^{11}$   \\
\hline

\multirow{3}{*}{Adult Census 15}                                   
& TabularQGAN                    & $0.913 \pm 0.008 $                      & 0.820                                 & 0.038            &  104 \\ 

& CTGAN                           & $0.813 \pm 0.008$                               & 0.925                                 & 0.117            & 131,072 \\ 
& CopulaGAN                       & $0.799 \pm 0.005$                               & 0.913                                 & 0.096            &   60 \\ 
& VAE-GMM                       & \bm{$0.979 \pm 0.004$}                               & 0.928                                 & 0.028          &  702   \\
& be-GReaT LLM                        & $0.899 \pm 0.015$                               & 0.960                                 & 0.037           &  $10^9-10^{11}$   \\\hline

\multirow{3}{*}{MIMIC 10}                                   
& TabularQGAN                    & \bm{$0.977 \pm 0.006$}                      & 0.973                                 & 0.006            & 88   \\ 

& CTGAN                           & $0.865 \pm 0.008$                               & 0.984                                 & 0.068            & 65,536 \\ 
& CopulaGAN                       & $0.863 \pm 0.005$                               & 0.981                                 & 0.062            &   131,072 \\
& VAE-GMM                       & $0.947 \pm 0.008$                              & 0.917                                 & 0.009           &  1,620  \\
& be-GReaT LLM                        & $0.928 \pm 0.017$                               & 0.987                                 & 0.234           &  $10^9-10^{11}$   \\\hline

\multirow{3}{*}{MIMIC 15} 
& TabularQGAN                    & \bm{$0.954 \pm 0.005$}                      & 0.784                                 & 0.133            & 37   \\ 

& CTGAN                           & $0.918 \pm 0.003$                               & 0.770                                 & 0.107            & 262,144 \\ 
& CopulaGAN                       & $0.912 \pm 0.005$                               & 0.757                                 & 0.118            &  131,072  \\ 
& VAE-GMM                       & $0.918 \pm 0.013$                               & 0.407                                 & 0.071           &  4,210   \\
& be-GReaT LLM                        & $0.880 \pm 0.006$                               & 0.969                                 & 0.267           &  $10^9-10^{11}$   \\\hline

\end{tabular}
}
\end{table}

The second column shows the results of the overlap fraction, which measures the number of unique rows in the synthetic data that are not present in the training data. Due to computational constraints, only a subsample of hyperparameter configurations is evaluated, namely the best and worst 10 models for each data and model type pair. If the overlap is one, it implies that no novel samples are generated. Our results show that each model does produce some novel samples, meaning they are not purely reproducing the training data set.

The number of parameters for the optimal configuration of the model varies across models and data sets, but for all but one data set, the classical model has far more parameters. Parameter counts for the quantum models are found by counting the number of tunable gates in the ansatz and multiplying by the optimally found number of layers. For the classical models, the number of parameters in each layer of the network is summed. The values reported in the last column represent the parameter count for the Boolean model configuration. Although the non-Boolean design employs fewer parameters, the difference is minimal. Hence, only the Boolean parameter counts are presented to maintain consistency with the results reported for other models. An analysis of the effect of circuit depth on model performance can be found in Appendix \ref{sec:circuit_depth_plots}.

In an attempt to find a more qubit-efficient alternative to one-hot encoding, we introduce a Unique-Row-Index encoding and train a generator composed of a single numerical register to reproduce the distribution of row indices (see Appendix~\ref{sec:unique-row-encoding-example}). However, the performance of this approach is significantly lower than that achieved with the proposed one-hot encoding using Givens rotations. This suggests that a single numerical register is not a suitable circuit design for generating samples with categorical features.

% \usepackage{multirow}

% \begin{figure}
% \begin{subfigure}[h]{0.5\linewidth}
% \includegraphics[width=1.1\linewidth]{figures/quantum_circuit_depth.pdf}
% \caption{}
% \label{fig:quantum_circuit_depth} 
% \end{subfigure}
% \hfill
% \begin{subfigure}[h]{0.5\linewidth}
% \includegraphics[width=1.1\linewidth]{figures/classical_circuit_depth.pdf}
% \label{fig:classical_circuit_depth}
% \caption{}
% \end{subfigure}%
% \label{fig:circuit_depth}
% \caption{(a) Effect of circuit depth on performance for quantum models. 
%  (b) Effect of number of layers on performance for classical models. Only data from  the boolean data encoding is included for both plots.}
% \end{figure}

% To be mentioned in this section
% - We confirmed that the model can learn a ground truth, however the limitation in qubit number did not allow us to reach the regime in which the model generates meaningful novel for both numerical and one-hot encoded categorical data sample

\section{Conclusion, Limitations, and Outlook}  
\label{sec:conclusion}
In this work, we introduce an adaptation of a quantum GAN model for tabular data. It utilizes a novel flexible encoding protocol and circuit ansatz to account for both numerical and one-hot encoded categorical data.  
We note that all presented results are obtained via a noiseless simulator on small feature subsets and constitute a proof-of-concept demonstration. In our experiments, the TabularQGAN model outperformed classical GAN models with respect to the overall similarity score and was comparable to a tabular specific variational auto encoder model and an LLM model on the datasets under consideration. With regard to the downstream score metric, we found that the VAE-GMM model and the TabularQGAN models performed the best, with the VAE-GMM having the lowest downstream score for all the data sets except MIMIC-10, where TabularQGAN had the lowest value. For the overlap metric there was no single model with the highest or lowest value across different data sets. Additionally, a high or low the overlap metric might be beneficial or detrimental to a model, depending on the intended use of the synthetic data, as noted in Section \ref{sec:metrics}.   
These different metrics were chosen to evaluate statistical similarity, novelty, and downstream performance on a supervised learning task; our results show that relative performance between models and datasets is not uniform, confirming the need to compare models carefully in a use case specific context. The quantum architecture has significantly fewer parameters than its classical counterparts. Training well-performing models for large-scale real-world applications can require expensive and energy-intensive computation. In this regime, the parameter compression provided by quantum models may reduce computational resources in future hardware settings.

While the proposed TabularQGAN demonstrates promising performance, further investigation will be valuable for understanding the behaviour of the model as the number of qubits increases. In the present study, the evaluation focused on reduced feature subsets of MIMIC-III and Adult Census datasets, enabling a controlled analysis within current computational constraints while still capturing representative characteristics of the underlying data distributions. This design choice reflects the practical constraints of simulating larger quantum systems on classical hardware, as training on current quantum devices remains costly and subject to noise.

The challenge of scaling variational quantum models to larger qubit counts has been discussed in the literature, notably in the context of the barren plateau phenomenon \cite{mcclean2018barren}. Although certain quantum architectures have been shown to mitigate such effects \cite{schatzki2024theoretical}, it remains an open question whether variational quantum models can scale to large numbers of qubits while avoiding classical simulability\cite{bermejo2024quantum, Cerezo2023DoesPA}. 

In future work, extending the evaluation to a broader range of datasets with higher-dimensional feature spaces would further strengthen the proof-of-concept for the proposed approach. In addition, training and sampling at scale on quantum hardware would provide valuable insights into the impact of device noise on sample quality and help clarify the potential performance gains as quantum hardware continues to improve in size and fidelity. Furthermore, extending the evaluation with a wider range of classical baselines, including diffusion-based methods such as TabDDPM\cite{tabDDPM}, TabSyn\cite{tabSyn}, and TabDiff\cite{tabDiff}, which currently represent strong baselines on tabular benchmarks, remains an important direction for future work. The competitive performance of TabularQGAN should be interpreted within the scope of the models and dataset configurations considered. Finally, while this work explores only two quantum circuit variants, further investigation into alternative ansatz designs and data encoding schemes may yield additional performance improvements. 

% Conclusion(1/2 page)
%    Relevance of our research and model
%    Potential Future work

{
\small
\section{Data Availability}
This manuscript uses open-source Adult Census \cite{adult_census} and MIMIC-III \cite{johnson2016mimic} datasets. A training and data use agreement is required to gain access to the MIMIC-III dataset \cite{mimic_db}.
\section{Code Availability}
The source code of TabularQGAN and the benchmarks is available at \href{https://zenodo.org/records/21264099}{zenodo.org/records/21264099}.

\bibliography{bibtex/references}

\section{Acknowledgements}

The authors would like to thank the other members of the QUTAC consortium. C.J. would also like to thank Abhishek Awasthi and Davide Vodola for their assistance with setting up HPC experiments. 
\section{Funding}
Each author is funded by their respective affiliated organization. There is no external funding provided to this project. While some organizations are members of Quantum Technology and Application Consortium (QUTAC), this membership does not entail financial support for this work as the consortium is based on in-kind contributions.

\section{Competing Interests}
The authors declare no competing interests.
\section{Author Contributions}

P.B, C.J and A.V conceived the study. P.B, C.J implemented the models and conducted the experiments. P.B, C.J, L.D contributed to data analysis and benchmarking against classical baselines. P.B, C.J, and L.D. interpreted the results and wrote the manuscript. P.B developed the theoretical framework in the paper. P.B, C.J and L.D contributed to the visualization of results. All authors reviewed and approved the final manuscript

\section{Figure Legends}

Figure 1: Non-Boolean circuit design (a) and Boolean circuit design (b) for a [n5,c3,c2] register. The model layer can be repeated $d$ times to obtain a depth-$d$ circuit. All qubits are measured in the computational basis to obtain a bitstring generated by the model. Here, the Boolean circuit design saves one qubit by treating the two-category feature as a Boolean variable and merging it into the numerical register.
\\
\newline
Figure 2: Schematic diagram of TabularQGAN training. In Step 1, either a batch of training data or a batch of synthetic samples (obtained from single-shot measurements) is fed to the discriminator. In Step 2, the discriminator attempts to distinguish between real and fake samples, and its parameters $\phi$ are updated based on the gradient of the discriminator loss~$L_{D}$. In Step 3, a sample is generated for each parameter shift, and the discriminator with fixed parameters $\phi$ is used to compute the gradient of the parameters according to the parameter-shift rule. In Step 4, the generator parameters~$\theta$ are updated based on their gradient.
\newline
\\
Table 1: Overview of datasets used in experiments, including the number of qubits, numerical and categorical features.
\newline
\\
Figure 3: Plot of the overall metric for each hyperparameter configuration for each dataset. The spread of the points within each bar is artificially added to improve visibility. 
\newline
\\
Figure 4: Plot showing the overlap fraction metric for (a) GAN type models and (b) VAE and LLM models. As explained in the text only a selected subset of the data is sampled.
\newline
\\
Table 2: Best performing models with respect to the overall metric for each data set and model type. The number of parameters is stated for the best-performing hyperparameter configuration. The parameter count of OpenAI's GPT-o4-mini, the base model of be-GReaT, is not disclosed, but we report a typical range for modern LLMs \cite{brown2020languagemodelsfewshotlearners}.
\newline
\\
Figure 5:  Downstream Score against Overall Metric for predicting categorical Variables for GANs (a) and VAE and LLM models (c). 
 Downstream Score against Overall Metric for predicting numeric Variables. for GANs (b)  and VAE and LLM model (d).   Only a selected subset of the data is sampled. 
}
\appendix
\clearpage % Removes first small part of appendix from references page
\begin{appendices}
\section{Appendices}

\subsection{Example of Tabular Encoding} \label{sec:encoding_example}
In Section \ref{sec:encoding}, we introduce a protocol for encoding classical tabular data into quantum states. Here we present an example of that encoding for the Adult Census 10 dataset with three features, age, income, and education. Age is a numerical feature, encoded with 5 qubits.  
\\
{Age} (numerical, \(N_{\rm age}=5\) qubits \(\rightarrow 2^5=32\) bins).
The width of each bin is calculated from the minimum and maximum values from the training data:
\begin{equation}
    W_{\text{age}} = (\{x_{\text{age}}\}_{\text{max}} - \{x_{\text{age}}\}_{\text{min}} )~/~\text{bin count}
\end{equation}
where $W_{\text{age}}$ is the width of each bin and $x_{\text{age}}$ is a vector of the age feature of the training data. The bin number is a rounded value of $x_{\text{age}} - \{x_{\text{age}}\}_{\text{min}} / W_{\text{age}}$. For example:

    \[
      \text{Age}=19 \;\mapsto\; \text{bin }2 \;\mapsto\; \lvert 00010 \rangle
    \]
    
Income is a (binary) variable with options  "<=50K" and ">50K". It requires one qubit for Boolean: \(\lvert0\rangle\) (<=50K), \(\lvert1\rangle\) (>50K) and 2-qubits for one-hot encoding: \(\lvert10\rangle\) (<=50K), \(\lvert01\rangle\) (>50K)

{Workclass is a categorical variable with four options } (4 classes, one-hot): 
\begin{align*}
    \centering
&(\text{empl-unknown}, \lvert1000\rangle),\; (\text{govt-employed}, \lvert0100\rangle) \\ &(\text{self-employed} ,\lvert0010\rangle),\; (\text{unemployed}, \lvert0001\rangle )  
\end{align*}
A single row is represented as: \\
\noindent{Encoding with Boolean design (10 qubits):} \[
  \{19,\;\text{"<=50K"},\;\text{govt-employed}\}
  \;\mapsto\;
  \lvert\,00010\,\|\;0\;\|\;0100\rangle
  = \lvert0001000100\rangle
\]

\noindent{Encoding with non-Boolean design (11 qubits):}
\[
   \{19,\;\text{"<=50K"},\;\text{govt-employed}\}
  \;\mapsto\;
  \lvert\,00010\,\|\;10\;\|\;0100\rangle
  = \lvert00010100100\rangle.
\]

\subsection{Example of Feature-Value Combination Index Encoding (Failure Case)} \label{sec:unique-row-encoding-example}

The number of qubits required to implement the one-hot encoding introduced in Section \ref{sec:encoding} scales linearly with the number of categories in one feature. It is natural to ask if a more qubit-efficient encoding is possible, while maintaining similarly high benchmarking results.

Consider the following encoding: Assign an index to every combination of feature values and encode this index as a binary number. We illustrate this encoding with the example from Appendix \ref{sec:encoding_example}. The set of all combinations of feature values is given by
\begin{equation}
    \mathcal{S} = \{ 0, \dots,  2^5 -1\} \times \{ \text{Income1}, \text{Income2} \} \times \{ \text{Workclass1}, \dots, \text{Workclass4} \}~,
\end{equation}
where $\times$ denotes the Cartesian product of two discrete sets. We assign an index to each of the $|\mathcal{S}| = 256 $ elements of the set and encode the index in binary using $\log_2 |\mathcal{S}| = 8$ qubits. This allows us to represent all elements of $\mathcal{S}$ using a single numerical register of 8 qubits. We train the circuit to generate indices that follow the underlying distribution in the training data and note that the generated indices can be decoded by directly accessing the corresponding element in $\mathcal{S}$.

\begin{figure}
    \centering
    \includegraphics[width=0.7\linewidth]{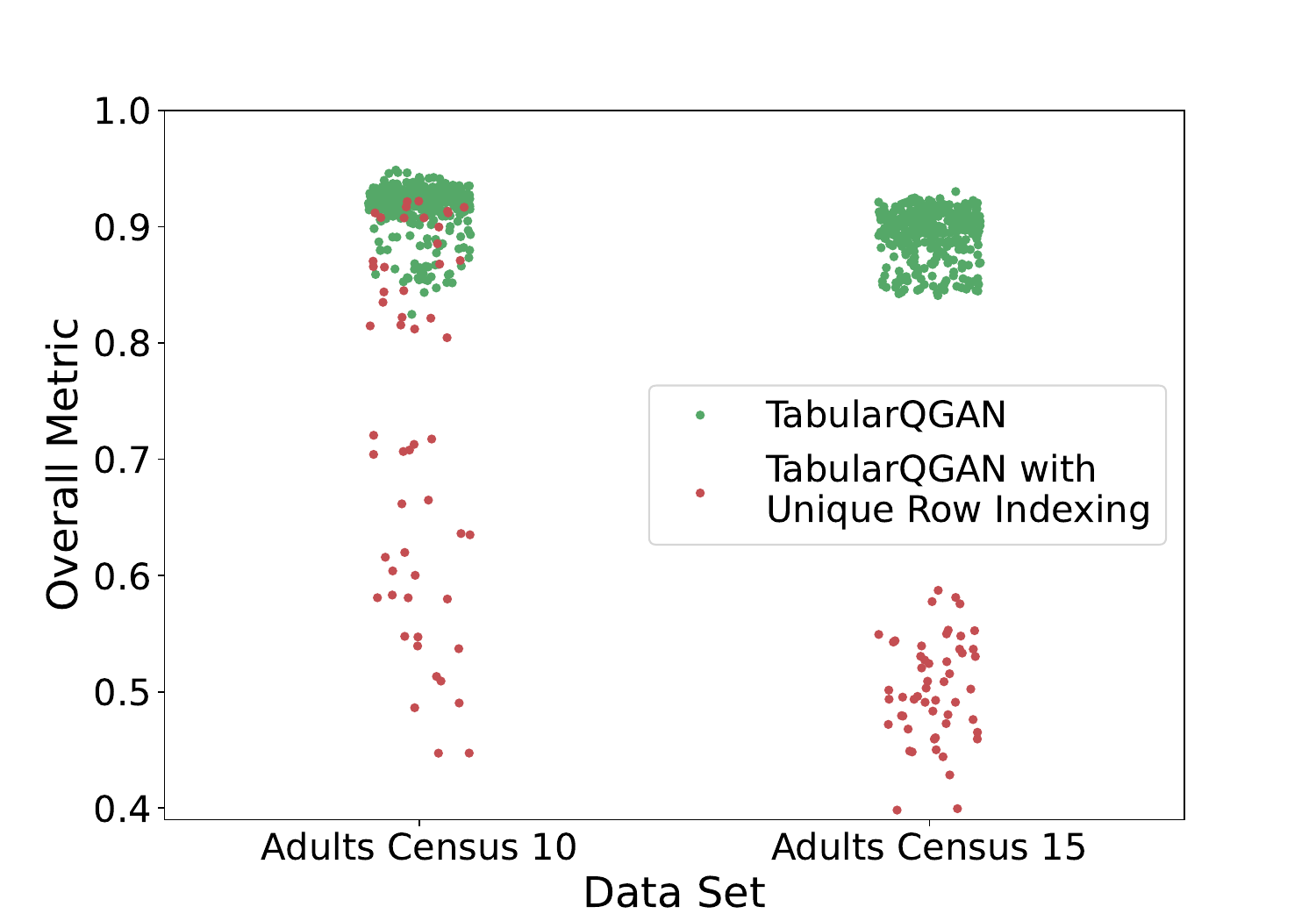}
    \caption{Comparison of the overall metric for each hyperparameter configuration on the Adult Census 10 and Adult Census 15 datasets, using the Unique-Row-Index encoding and a single numerical register for index generation. The spread of points within each bar has been added to improve data visibility. It is evident that the performance of the TabularQGAN model is significantly lower when a Unique-Row-Index encoding is used instead of the proposed one-hot encoding.}
    \label{fig:unique-row-overall-metric}
\end{figure}

Using this encoding, the model is trained on the same hyperparameter ranges described in Section~\ref{sec:hyper_param_opt}, applied to both the Adult Census 10 and Adult Census 15 datasets. While scores above 0.9 are achieved on the Adult Census 10 dataset for a few hyperparameter configurations, the performance on the Adult Census 15 dataset is significantly lower (Figure \ref{fig:unique-row-overall-metric}). This suggests that a single numerical register is not a suitable circuit design for generating samples with categorical features.

\subsection{Overall Metric Details} \label{sec:sd_definition_score}
The overall metric from the SDMetrics library \cite{sdmetrics}, called the overall similarity score is an average over two components, the first is a column-wise metric, $S_{\text{shape}}$, and the second is a pairwise metric over each pair of columns $S_{\text{pair}}$.

The column shape metric is given by $S_{\text{shape}} = \frac{1}{C} \sum_{i=1}^{C} s_i$ where $C$ is the number of columns and $s_i$ is the Kolmogorov–Smirnov complement for numerical columns ($s_i = 1 - \mathrm{KS}(P_i, Q_i)$) and the Total Variational Distance Complement for categorical columns ($s_i = 1 - \frac{1}{2} \sum_k |P_i(k) - Q_i(k)|$). Here, $Q_i$ is a vector of the values of column~$i$ from the training data, and $P_i$ is the equivalent for the synthetic data. For the categorical data, $Q_i(k)$ ($P_i(k)$) is the count of the instances of the category $k$ in column  $Q_i$ ($P_i(k)$).  \\

 The pair-wise metric is given by $S_{\text{pair}}~=~\frac{1}{\binom{C}{2}}~\sum_{i<j}~t_{ij}$, for columns  $i$ and $j$. 
 For numerical data, $t_{ij} =  1 - (|P^S_{ij} - P^R_{ij}| /2)$  where  $P^S_{ij}$ and $P^R_{ij}$ are the Pearson correlation coefficients for the synthetic and real data, respectively. For categorical data (or a mixed pair of categorical and numerical data), $t_{ij} = 1 - \frac{1}{2} \sum_{\alpha \in A_{i}} \sum_{\beta \in B_{j}}  |  F^S_{\alpha \beta }   - F^R _{\alpha \beta }  | $. This is the contingency similarity where $\alpha$ denotes each of the categories of column $A_i$, and  $ F^S_{\alpha \beta } $ is the frequency of the category values  $\alpha$  and  $\beta$ for the synthetic data. 
Each of these metrics is normalized such that they are between $[0,1]$ with 1 being perfect similarity.

The final overall metric is an average of the two components. $S_{\text{overall}} = \frac{1}{2}(S_{\text{shape}} + S_{\text{pair}})$. 

\subsection{Data Preprocessing and Binning} \label{sec:preprocessing}
As described in the main text, both the MIMIC and Adult Census datasets were preprocessed for each of the different model types. Here we list those steps. 
\subsubsection{TabularQGAN model}
 The raw MIMIC data consist of many tables but for this work, we combined the admissions, patients and ICU stays tables.
\begin{itemize}
    \item The  age was constructed from the date of birth and admission date columns, filtered to remove unphysical ages (a known artifact due to the anonymization of the dataset) then binned into 64 bins  (corresponding to 6 qubits) and converted into bitstrings. 
\item Admission Time was similarly transformed except with 16 bins, corresponding to 4 qubits.
\item The Gender feature already existed in the dataset and was mapped to a bitstring (either one or two bits depending on whether we used Boolean encoding). 
\item The raw admission type column had three options, which were one-hot encoded to a bitstring of length three. 
\item The overall qubit breakdown for MIMIC 10: AGE: 6, GENDER: 1, ADMISSION TYPE: 3
\item The overall qubit breakdown for MIMIC 15: AGE: 6, ADMISSION TIME: 4,  GENDER: 1, ADMISSION TYPE: 3
\end{itemize}
For the Adult Census Data:
\begin{itemize}
    \item The Age feature, constructed by taking the raw age as an integer, was binned into 32 bins and converted to a length-5 bitstring 
    \item The Income feature was made by mapping the raw categorical feature  of above or below 50K into a Boolean value. 
    \item The Workclass feature was created by mapping the raw workclass categorical feature to a simplified set of 4 options which was one-hot encoded to a bitstring of length 4. 
    \item  The Education feature was created by mapping the raw education categorical feature to a simplified set of 6 options which was one-hot encoded to a bitstring of length 6. 
    \item The overall qubit breakdown for Adult Census 10:  AGE: 5, INCOME: 1, WORKCLASS: 4
    \item The overall qubit breakdown for Adult Census 15: AGE: 5, WORKCLASS: 4, EDUCATION: 6
\end{itemize}
\subsubsection{Classical Models}
For the classical models, the same data preprocessing transformations were applied to the data as for the TabularQGAN model, with the exception that the data was not converted into bitstrings but instead left as a combination of numbers and text. Thus, the same data filtering, binning and encoding steps were applied, and only the final conversion to a bitstring is different. For example, a classical model would see a sample of the Adult Census 10 dataset as \[
  \{19,\;\text{"<=50K"},\;\text{govt-employed}\}
  \]
and the TabularQGAN Model would see:
  \[
  \lvert\,00010\,\|\;0\;\|\;0100\rangle
  = \lvert0001000100\rangle
\]
\subsection{Hyperparameter Search Space}
The Table \ref{tab:hyperparameter_grid} lists all the different hyperparameter settings  optimized across the different model types. 
\begin{table}[h!]
\centering
\caption{Hyperparameter settings across model types}
\label{tab:hyperparameter_grid}
\subcaption{Hyperparameter settings for the GAN model types }

\vspace{0.6em} 
{\scriptsize
\begin{tabular}{lllllll}
\hline
\textbf{Model Type} & \begin{tabular}[c]{@{}l@{}}\textbf{Circuit} \\ \textbf{Depth}\end{tabular} & \begin{tabular}[c]{@{}l@{}} \textbf{Batch} \\ \textbf{Size Fraction} \end{tabular} & \begin{tabular}[c]{@{}l@{}}\textbf{Learning Rate} \\ \textbf{Discriminator}\end{tabular} & \begin{tabular}[c]{@{}l@{}}\textbf{Learning Rate} \\ \textbf{Generator}\end{tabular} & \textbf{Layer Width} & \begin{tabular}[c]{@{}l@{}}\textbf{Total} \\ \textbf{Epochs}\end{tabular} \\ \hline
TabularQGAN & 1, 2, 3, 4 & 0.1, 0.2 & 0.05, 0.1, 0.2 & 0.05, 0.1, 0.2 & \hspace{2em}- & 300, 600 \\ 
CTGAN & 1, 2, 3, 4 & 0.1, 0.2 & 0.001, 0.01, 0.05 & 0.001, 0.01, 0.05 & \hspace{2em} Data Width, 256 & 1500 \\ 
CopulaGAN & 1, 2, 3, 4 & 0.1, 0.2  & 0.001, 0.01, 0.05 & 0.001, 0.01, 0.05 & \hspace{2em} Data Width, 256 & 1500 \\ \hline
\end{tabular}
}
\qquad
\subcaption{Hyperparameter settings for the VAE-GMM Model }
{\scriptsize
\begin{tabular}{llllll}
\hline
\textbf{Model Type} & \begin{tabular}[c]{@{}l@{}}\textbf{Hidden} \\ \textbf{Dimension}\end{tabular} & \begin{tabular}[c]{@{}l@{}} \textbf{Batch} \\ \textbf{Size} \end{tabular} & \begin{tabular}[c]{@{}l@{}}\textbf{Learning} \\ \textbf{Rate}\end{tabular} & \begin{tabular}[c]{@{}l@{}}\textbf{Latent} \\ \textbf{Dimension}\end{tabular} & \begin{tabular}[c]{@{}l@{}}\textbf{Number} \\ \textbf{Epochs}\end{tabular} \\ \hline
VAE-GMM & 50, 100, 150 & 500, 1000 & 0.0001, 0.001, 0.01 & 2, 5, 10, 20  & 200 \\ 
 \hline
\end{tabular}
}
\subcaption{Hyperparameter settings for the be-GReaT LLM Model }
    {\scriptsize  
    \begin{tabular}{lllllllll}
    \hline
    \textbf{Model Type} & \begin{tabular}[c]{@{}l@{}} \textbf{Temp.} \end{tabular} & \begin{tabular}[c]{@{}l@{}} \textbf{Warm-up} \\ \textbf{ratio} \end{tabular} & \begin{tabular}[c]{@{}l@{}} \textbf{Sample max} \\ \textbf{length} \end{tabular}  &  \begin{tabular}[c]{@{}l@{}} \textbf{Sample k} \end{tabular}  &  \begin{tabular}[c]{@{}l@{}} \textbf{Batch} \\ \textbf{Size} \end{tabular} & \begin{tabular}[c]{@{}l@{}} \textbf{Learning} \\ \textbf{rate}\end{tabular}  &  \begin{tabular}[c]{@{}l@{}} \textbf{Weight} \\ \textbf{decay} \end{tabular} & \begin{tabular}[c]{@{}l@{}} \textbf{Number} \\ \textbf{Epochs} \end{tabular} \\ \hline
         be-GReaT   & 0.3, 0.5, 0.7                 & 0, 0.1                     & 50, 100     &      50      &    8      &  \begin{tabular}[c]{@{}l@{}} 0.00001,\\0.00003,\\ 0.00007 \end{tabular}                                        & 0, 0.01     & 1              
                          \\   \hline
    \end{tabular}
    }
\end{table}

\subsection{Optimal Hyperparameter Configurations} \label{sec:opt_hyperparameter}

Here we show the hyperparameter configurations associated with our best-found models mentioned in Table~\ref{tab:best_found_models}. Each configuration was selected by taking the model instance with the maximum overall metric for each dataset. For TabularQGAN, we found that the deeper circuit depths were optimal, and for the classical models, a lower number of layers was better.
\begin{table}[h]
    \centering
    \caption{Best found hyperparameter configurations for each model and data type}
    \label{tab:best_found_models}
    \subcaption{ GAN Models}
    \vspace{0.6em}
    {\scriptsize  
    \begin{tabular}{llllllll}
    \hline
    \begin{tabular}[c]{@{}l@{}} \textbf{Dataset} \\ \textbf{Name} \end{tabular} & \textbf{Model} & \begin{tabular}[c]{@{}l@{}}  \textbf{Batch} \\ \textbf{Size Fraction} \end{tabular} & \begin{tabular}[c]{@{}l@{}} \textbf{Circuit} \\ \textbf{Depth} \end{tabular} & \begin{tabular}[c]{@{}l@{}} \textbf{LR} \\ \textbf{Gen} \end{tabular}  &  \begin{tabular}[c]{@{}l@{}} \textbf{LR} \\ \textbf{Discrim} \end{tabular} &  \begin{tabular}[l]{@{}l@{}} \textbf{Layer} \\ \textbf{Width} \end{tabular} &  \begin{tabular}[c]{@{}l@{}} \textbf{Number} \\ \textbf{Epochs}   \end{tabular}  \\  \hline
          & TabularQGAN   & 0.2                   & 4                      & 0.200                            & 0.050                              & N/A                 & 1991                 \\
     \begin{tabular}[l]{@{}l@{}} Adult  \\  Census 10\end{tabular}  & CTGAN          & 0.2                   & 2                      & 0.001                            & 0.050                              & 256                 & 1500                 \\ 
          & CopulaGAN      & 0.2                   & 1                      & 0.001                            & 0.010                              & \begin{tabular}[l]{@{}l@{}} Data  \\  Width\end{tabular}                  & 1500                    \\   \hline
     & TabularQGAN   & 0.1                   & 4                      & 0.200                            & 0.100                              & N/A                 & 2114                   \\    

         \begin{tabular}[l]{@{}l@{}} Adult  \\  Census 15\end{tabular}  & CTGAN          & 0.2                   & 2                      & 0.050                            & 0.001                              & 256                 & 1500                  \\ 
         & CopulaGAN      & 0.1                   & 1                      & 0.001                            & 0.050                              & \begin{tabular}[l]{@{}l@{}} Data  \\  Width\end{tabular}          & 1500                     \\ 
     \hline
            & TabularQGAN   & 0.2                   & 4                      & 0.100                            & 0.050                              & N/A                 & 2450    \\
         \begin{tabular}[l]{@{}l@{}} MIMIC 10  \\ \:\end{tabular}           & CTGAN          & 0.1                   & 1                      & 0.001                            & 0.010                              & 256                 & 1500                 \\ 
                 & CopulaGAN      & 0.1                   & 2                      & 0.001                            & 0.010                              & 256                 & 1500            \\ 
                      \hline

        & TabularQGAN   & 0.1                   & 4                      & 0.200                            & 0.100                              & N/A          &  1778  \\
           \begin{tabular}[l]{@{}l@{}} MIMIC 15   \\ \: \end{tabular}         & CTGAN          & 0.1                   & 4                      & 0.001                            & 0.010                              & 256                 & 1500               \\
               & CopulaGAN      & 0.1                   & 2                      & 0.001                            & 0.010                              & 256                 & 1500                
                         \\ \hline
    \end{tabular}
    }
    \subcaption{VAE model}
    {\scriptsize  
    \begin{tabular}{lllllll}
    \hline
    \begin{tabular}[c]{@{}l@{}} \textbf{Dataset} \\ \textbf{Name} \end{tabular} & \textbf{Model} & \begin{tabular}[c]{@{}l@{}} \textbf{Hidden} \\ \textbf{Dimension} \end{tabular} & \begin{tabular}[c]{@{}l@{}} \textbf{Batch} \\ \textbf{Size} \end{tabular} & \begin{tabular}[c]{@{}l@{}} \textbf{Learning} \\ \textbf{Rate} \end{tabular}  &  \begin{tabular}[c]{@{}l@{}} \textbf{Latent} \\ \textbf{Dimension} \end{tabular}  &  \begin{tabular}[c]{@{}l@{}} \textbf{Number} \\ \textbf{Epochs} \end{tabular} \\ \hline
    \begin{tabular}[l]{@{}l@{}} Adult  \\  Census 10\end{tabular}
          & VAE-GMM   & 150                 & 500                     & 0.0001                           & 2                                         & 200                   
                          \\   \hline
    \begin{tabular}[l]{@{}l@{}} Adult  \\  Census 15\end{tabular}
          & VAE-GMM   & 50                 & 500                     & 0.0001                           & 2                                         & 200                   
                          \\   \hline
    MIMIC 10 
          & VAE-GMM   & 50                 & 1000                     & 0.0001                           & 20                                         & 200                   
                          \\   \hline
    MIMIC 15
          & VAE-GMM   & 150                 & 500                     & 0.01                           & 10                                         & 200                   
                          \\   \hline
    \end{tabular}
    }
    \subcaption{LLM model}
    {\scriptsize  
    \begin{tabular}{llllllllll}
    \hline
    \begin{tabular}[c]{@{}l@{}} \textbf{Dataset} \\ \textbf{Name} \end{tabular} & \textbf{Model} & \begin{tabular}[c]{@{}l@{}} \textbf{Temp.} \end{tabular} & \begin{tabular}[c]{@{}l@{}} \textbf{Warm-up} \\ \textbf{ratio} \end{tabular} & \begin{tabular}[c]{@{}l@{}} \textbf{Sample max} \\ \textbf{length} \end{tabular}  &  \begin{tabular}[c]{@{}l@{}} \textbf{Sample k} \end{tabular}  &  \begin{tabular}[c]{@{}l@{}} \textbf{Batch} \\ \textbf{Size} \end{tabular} & \begin{tabular}[c]{@{}l@{}} \textbf{Learning} \\ \textbf{rate}\end{tabular}  &  \begin{tabular}[c]{@{}l@{}} \textbf{Weight} \\ \textbf{decay} \end{tabular} & \begin{tabular}[c]{@{}l@{}} \textbf{Number} \\ \textbf{Epochs} \end{tabular} \\ \hline
    \begin{tabular}[l]{@{}l@{}} Adult  \\  Census 10\end{tabular}
          & be-GReaT   & 0.7                 & 0-0.1                     & 50     &      50      &    8      &  0.00001                                        & 0.0     & 1              
                          \\   \hline
    \begin{tabular}[l]{@{}l@{}} Adult  \\  Census 15\end{tabular}
          & be-GReaT   & 0.7                 & 0-0.1                     & 100      &     50    &       8     & 0.00001                                         & 0.0      & 1            
                          \\   \hline
    MIMIC 10 
          & be-GReaT   & 0.7                 & 0-0.1                     & 100        &     50   &     8      & 0.00001                                         & 0.0       & 1            
                          \\   \hline
    MIMIC 15
          & be-GReaT   & 0.5                 & 0-0.1                     & 50          &    50    &     8    & 0.00003                                         & 0.0           & 1       
                          \\   \hline
    \end{tabular}
    }
\end{table}

\subsection{Effect of Different Circuit Encodings} \label{sec:circuit_encoding}
\begin{figure}[h!]
    \centering
    \includegraphics[width=0.7\linewidth]{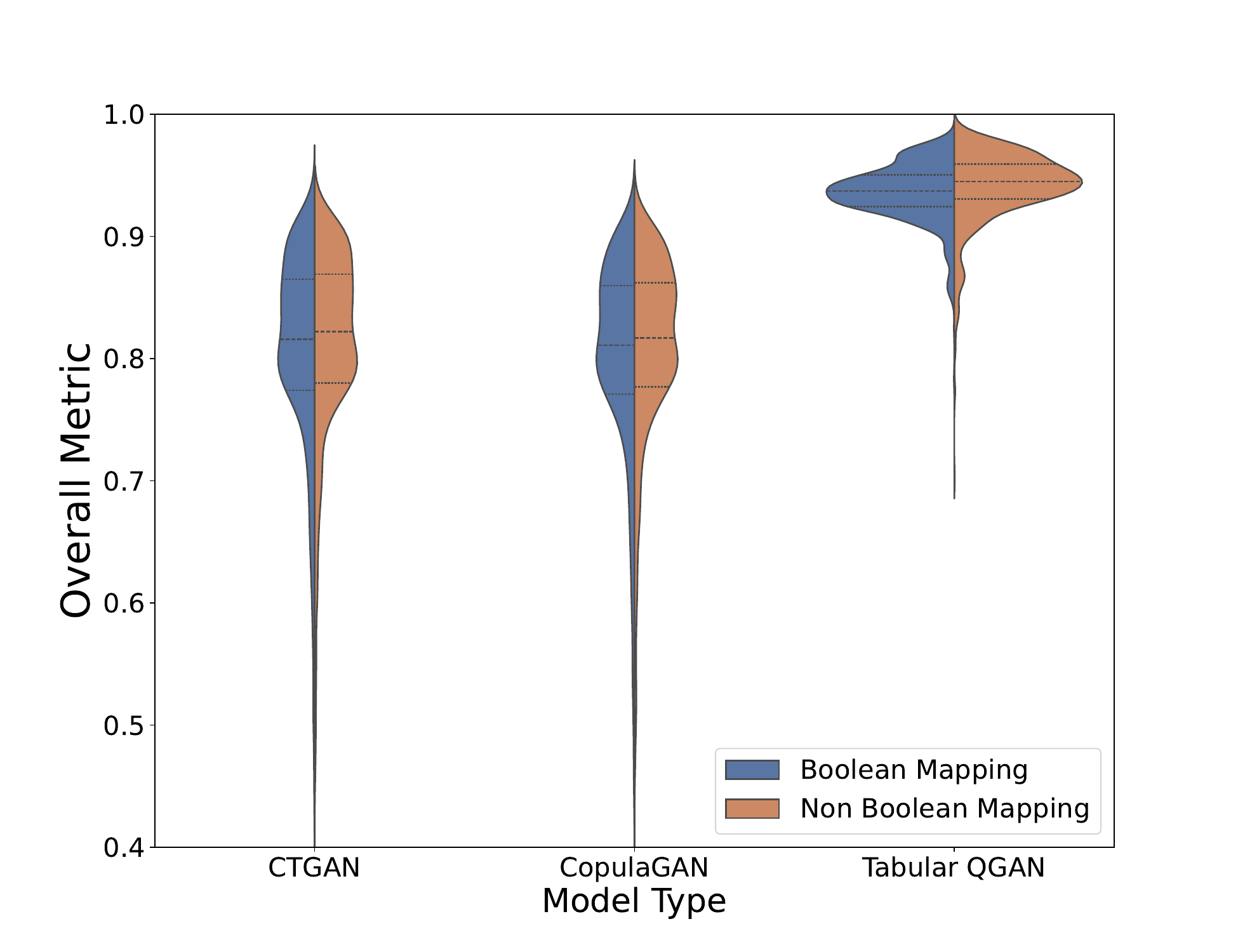}
    \caption{Plot showing the distribution of the overall metric value for each dataset with the two different encodings. Adult Census 15 is excluded as it does not contain any binary features. }
    \label{fig:boolean_circuit_eval}
\end{figure}

We explored two different ways of encoding Boolean variables for the GAN models, with one or two qubits, as described in Section \ref{sec:encoding}. We performed the hyperparameter optimization search for each different encoding, for each model type. Figure \ref{fig:boolean_circuit_eval} shows the distributions of the overall metric over the hyperparameter configurations. The distributions for the two different encoding types exhibit high similarity, indicating that the impact of encoding choice was minimal for the features and datasets considered. This limited effect is likely due to the datasets containing at most two Boolean features, which affects only one or two qubits under the current encoding method. However, as the number of Boolean features increases, the influence of encoding choice on performance may become more significant. 

\subsection{Effect of Circuit Depth on Performance} \label{sec:circuit_depth_plots}

\begin{figure}[h]
\centering
\begin{subfigure}{0.49\linewidth}
    \centering
    \includegraphics[trim={40pt 0pt 40pt 0pt},clip,width=\linewidth]{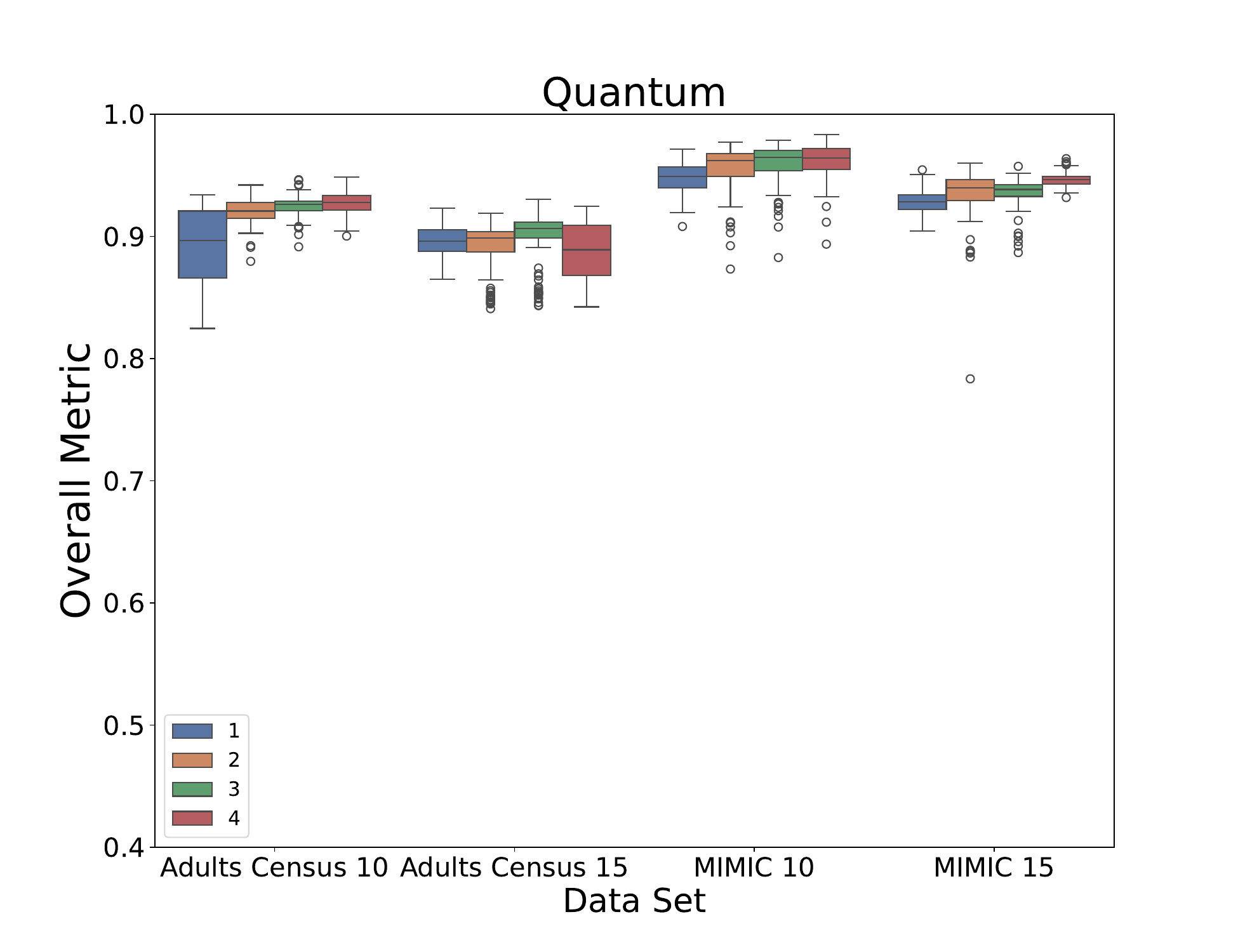}
    \caption{}
    \label{fig:quantum_circuit_depth_overall} 
\end{subfigure}
\hfill
\begin{subfigure}{0.49\linewidth}
    \centering
    \includegraphics[trim={40pt 0pt 40pt 0pt},clip,width=\linewidth]{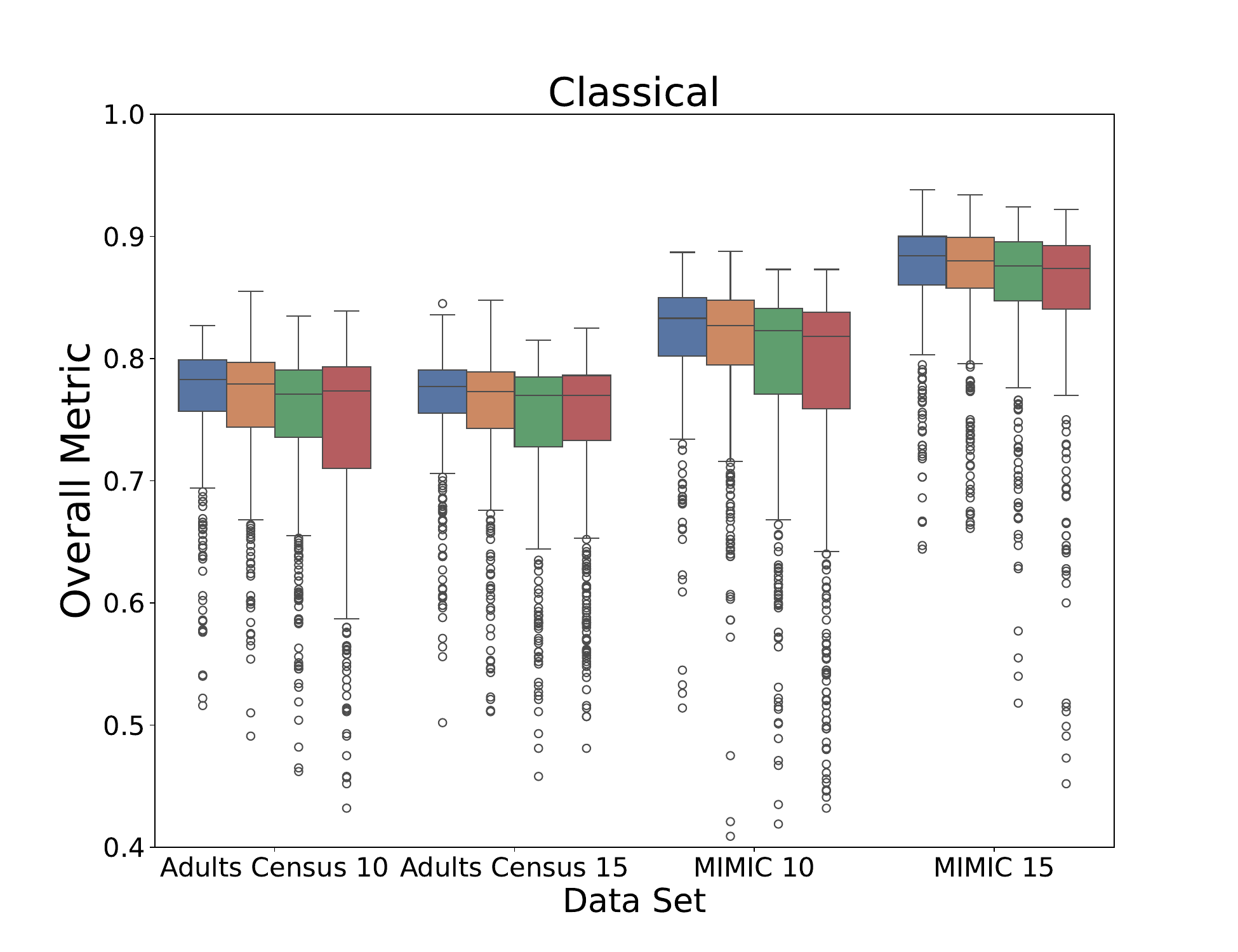}
    \caption{}
    \label{fig:classical_circuit_depth_overall}
\end{subfigure}
\caption{(a) Effect of circuit depth on performance for quantum models. 
(b) Effect of the number of layers on performance for classical models. Only data from the Boolean encoding is included for both plots.}
\label{fig:circuit_depth}
\end{figure} 
In Figure \ref{fig:circuit_depth}, the effect of circuit depth for our quantum model and number of layers for the classical GAN models is explored, with the results averaged over all other hyperparameter settings. Only the GAN type models are considered in this analysis as the VAE and LLM models do not have the circuit depth parameter.  For the TabularQGAN model (Figure \ref{fig:quantum_circuit_depth_overall}), an increasing circuit depth had a small performance improvement for all datasets except for the Adult Census 15 dataset, where there was little difference. The classical models (Figure \ref{fig:classical_circuit_depth_overall}) showed the opposite behavior: increasing the number of layers leads to worse performance. As the classical models had much larger parameter counts, we speculate that this may have been due to excessive overparameterization, which could lead to smaller gradients, slowing, or even stopping training.

\end{appendices}
\end{document}